%% file: main.tex
\documentclass[10pt,twocolumn,letterpaper]{article}

\usepackage[pagenumbers]{cvpr} 

\usepackage{graphicx}
\usepackage{amsmath}
\usepackage{amssymb}
\usepackage{booktabs}
\usepackage{siunitx}
\usepackage{cite}
\usepackage{ctable}
\usepackage{hhline}
\usepackage{booktabs} 
\usepackage{subcaption}
\usepackage{csquotes}
\usepackage{multirow}
\usepackage{pifont}
\usepackage[super]{nth}
\usepackage{mathtools,bbm}
\usepackage{bm}
\usepackage[warn]{textcomp}
\usepackage{wrapfig}

\newcolumntype{L}[1]{>{\raggedright\let\newline\\\arraybackslash\hspace{0pt}}m{#1}}
\newcolumntype{C}[1]{>{\centering\let\newline\\\arraybackslash\hspace{0pt}}m{#1}}
\newcolumntype{R}[1]{>{\raggedleft\let\newline\\\arraybackslash\hspace{0pt}}m{#1}}

\definecolor{cvprblue}{rgb}{0.21,0.49,0.74}
\usepackage[pagebackref,breaklinks,colorlinks,citecolor=cvprblue]{hyperref}

\def\confName{CVPR}
\def\confYear{2024}

\title{Authentic Hand Avatar from a Phone Scan via Universal Hand Model}

\author{
Gyeongsik Moon\\
\and
Weipeng Xu\\
\and
Rohan Joshi\\
\and
Chenglei Wu\\
\and
Takaaki Shiratori\\
\and
\vspace{-10mm}
\\
Codec Avatars Lab, Meta
\\
\url{https://mks0601.github.io/UHM}
}

\begin{document}

\twocolumn[{
\maketitle
\vspace{-2.5em}
\centerline{
\includegraphics[width=\linewidth,trim={4pt 4pt 4pt 4pt}]{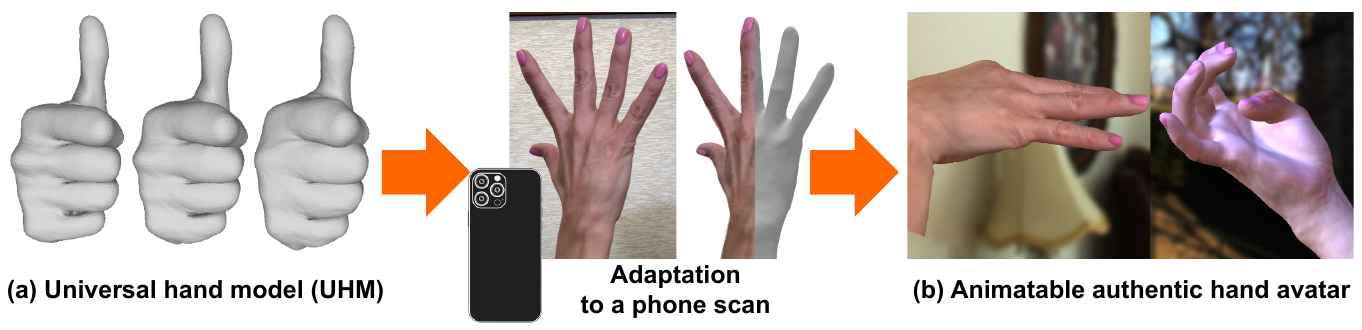}}
\vspace{-0.5em}
\captionof{figure}
{
We introduce (a) UHM, which can universally represent arbitrary IDs of hands at a high fidelity.
Our adaptation pipeline fits pre-trained UHM to a phone scan, which produces (b) an animatable authentic 3D hand avatar.
Images of (b) are rendered using our adapted hand avatar with the Phong reflection model and environment maps~\cite{gardner2017learning,hold2019deep}.
}
\label{fig:teaser}
\vspace{1em}
}]

\maketitle

\begin{abstract}
The authentic 3D hand avatar with every identifiable information, such as hand shapes and textures, is necessary for immersive experiences in AR/VR.
In this paper, we present a universal hand model (UHM), which 1) can universally represent high-fidelity 3D hand meshes of arbitrary identities (IDs) and 2) can be adapted to each person with a short phone scan for the authentic hand avatar.
For effective universal hand modeling, we perform tracking and modeling at the same time, while previous 3D hand models perform them separately.
The conventional separate pipeline suffers from the accumulated errors from the tracking stage, which cannot be recovered in the modeling stage.
On the other hand, ours does not suffer from the accumulated errors while having a much more concise overall pipeline.
We additionally introduce a novel image matching loss function to address a skin sliding during the tracking and modeling, while existing works have not focused on it much.
Finally, using learned priors from our UHM, we effectively adapt our UHM to each person's short phone scan for the authentic hand avatar.
\end{abstract}

\input{src/introduction}

\input{src/related_works}

\input{src/uhm}

\input{src/track_and_model}

\input{src/adaptation_to_a_phone_scan}

\input{src/experiment}

\input{src/conclusion}

\clearpage

\input{src_suppl/main_suppl}

\input{src_suppl/more_qualitative_results}
\input{src_suppl/more_ablation_studies}

\input{src_suppl/uhm_network_architectures_loss_functions}

\input{src_suppl/details_adaptation_to_a_phone_scan}

\input{src_suppl/our_datasets}
\input{src_suppl/experiment_details}
\input{src_suppl/failure_cases}

\clearpage

{
\small
\bibliographystyle{ieeenat_fullname}
\bibliography{bib}
}

\end{document}

%% file: src/introduction.tex
\section{Introduction}

We, humans, interact with the world through our hands.
We interact with other people with hand gestures, express our feelings through hand motions, and interact with objects with diverse hand poses.
The authentic 3D hand avatar with every identifiable information, including 3D hand shape and texture, is necessary for immersive experiences in AR/VR.

A 3D hand model is a function that produces a 3D hand from a 3D pose and identity (ID) latent code.
The pose represents 3D joint angles, and the ID latent code determines identifiable hand shape (\textit{e.g.}, thickness and size) in the zero pose or textures (\textit{e.g.}, skin color and fingernail polish).
Such two inputs (\textit{i.e.}, 3D pose and ID code) are used to drive pre-trained 3D hand models, where the 3D poses can be obtained from 3D hand pose estimators~\cite{ge20193d,choi2020p2m,moon2020i2l,lin2021end,lin2021mesh,moon2022hand4whole} and ID latent code can be obtained in a personalization stage~\cite{karunratanakul2023harp}.
Those two inputs are relatively affordable data from single or stereo camera setup of in-the-wild environment than 3D reconstruction~\cite{guo2019relightables}, which requires at least tens of cameras.
Hence, the 3D hand model is a core component of the 3D hand avatar.

We present a universal hand model (UHM), which 1) can universally represent high-fidelity 3D hand meshes of arbitrary IDs like Fig.~\ref{fig:teaser} (a) and 2) can be adapted to each person with a short phone scan for the authentic hand avatar like Fig.~\ref{fig:teaser} (b).
For the effective universal hand modeling, we perform the tracking and modeling at the same time, while existing 3D hand models~\cite{romero2017embodied,li2022nimble,potamias2023handy,corona2022lisa,chen2023handavatar,mundra2023livehand,iwase2023relightablehands} rely on a separate tracking and modeling pipeline.
Their tracking stage~\cite{amberg2007optimal,hirshberg2012coregistration} prepares target 3D meshes by non-rigidly aligning a template mesh to targets, such as 3D joint coordinates, 3D scans, masks, and images.
In this way, the tracking stage provides 3D meshes with a consistent topology across all captures.
Then, a modeling stage supervises 3D hand models with the tracked 3D meshes.
One of the limitations of such a conventional separated pipeline is that the tracking errors cannot be recovered in the modeling stage, which we call \emph{error accumulation problem}.
On the other hand, as our UHM performs the tracking and modeling at the same time in a single stage, it does not suffer from the error accumulation problem while the overall pipeline becomes much more concise.

We additionally propose an optical flow-based loss function to prevent skin sliding during the tracking and modeling, while existing 3D hand models have not focused on it much.
Most 3D hand models~\cite{romero2017embodied,li2022nimble,potamias2023handy} are simply trained by minimizing per-vertex distance against tracked 3D meshes, and the tracking~\cite{amberg2007optimal,hirshberg2012coregistration} is performed by minimizing iterative closest point (ICP) distance against 3D scans.
There could be a number of correspondences between 3D scans and 3D meshes from the 3D hand models as they do not share the same mesh topology.
Therefore, without proper objective functions, some vertices of the 3D hand models could \emph{slide} to semantically wrong positions.
For example, although a group of vertices is supposed to be consistently located at the thumbnail across all captures, due to the ambiguity of the ICP loss, they could be \emph{slid} to the below of the thumbnail.
To address this, we propose an image matching loss function, which minimizes the norm of the optical flow between our rendered images and captured images.
The optical flow provides image-level correspondences, especially useful for distinctive hand parts, such as fingernails and wrinkles on the palm.
As we use a deep optical flow estimation network~\cite{teed2020raft}, which can recognize contextual information of images, the optical flow provides semantically meaningful correspondences, while the ICP loss does not.

Most importantly, we introduce an effective pipeline for adapting our UHM to each person with a short phone scan, which gives the authentic hand avatar.
We found that existing works~\cite{karunratanakul2023harp} produce plausible outputs, but they lack authenticity, for example, slightly different 3D hand shapes from the target hand.
On the other hand, with the help of useful priors from the tracking and modeling stage, we successfully achieve highly authentic results.

Our contributions can be summarized as follows.
\begin{itemize}
\item We present UHM, a 3D hand model that can 1) universally represent high-fidelity 3D hand meshes of arbitrary IDs and 2) be adapted to each person with a short phone scan for the authentic 3D hand avatar.
\item UHM performs the tracking and modeling at the same time, while existing models perform them separately, to address the accumulated errors from the modeling stage.
\item We propose a novel image matching loss function to address the skin sliding problem during the tracking and modeling.
\item We propose an effective adaptation pipeline for the authentic hand avatar, which utilizes useful priors from the tracking and modeling stage.
\end{itemize}

%% file: src/related_works.tex
\section{Related works}

\noindent\textbf{3D hand models.}
Universal 3D hand modeling aims to train a 3D hand model that can universally represent 3D hands of arbitrary IDs.
MANO~\cite{romero2017embodied} is one of the pioneers in universal 3D hand modeling, and it is the most widely used one.
NIMBLE~\cite{li2022nimble} is a 3D hand model that consists of bones, muscles, and skin mesh.
LISA~\cite{corona2022lisa} is based on the implicit representation, motivated by neural radiance field~\cite{mildenhall2020nerf}.
Handy~\cite{potamias2023handy} is a high-fidelity 3D hand model that follows a formulation of MANO.
Due to the difficulty of universal modeling and collecting large-scale data from multiple IDs, there have been introduced several personalized 3D hand models.
Those personalized 3D hand models can only represent a single ID of the training set and cannot represent novel IDs.
DHM~\cite{moon2020deephandmesh} is a high-fidelity personalized 3D hand model.
LiveHand~\cite{mundra2023livehand} and HandAvatar~\cite{chen2023handavatar} are based on the implicit 3D representation of hands, inspired by neural radiance field~\cite{mildenhall2020nerf}.
RelightableHands~\cite{iwase2023relightablehands} is a relightable personalized 3D hand model.

Compared to the above 3D hand models, our UHM has three distinctive advantages.
First, UHM performs the tracking and modeling at the same time to address the \emph{error accumulation problem} from the tracking stage.
Second, we introduce a novel image matching loss function to address the skin sliding issue during the tracking and modeling.
Finally, ours can produce authentic hand avatar from a phone scan, while previous models~\cite{corona2022lisa,chen2023handavatar,mundra2023livehand} require accurate 3D keypoints and MANO registrations of capture studio datasets~\cite{Moon_2020_ECCV_InterHand2.6M}.
In addition, their texture modules produce images of studio space~\cite{Moon_2020_ECCV_InterHand2.6M,moon2020deephandmesh}, which has a big appearance gap from phone capture images.
The texture module of Handy~\cite{potamias2023handy} fails to replicate person-specific details, such as fingernail polish and tattoos, due to the limited expressiveness of their latent space.

\noindent\textbf{3D hand avatar from a phone scan.}
Creating a 3D hand avatar from a short phone scan has been started to be studied recently.
The 3D hand avatar should 1) be personalized to a target person with authenticity including 3D hand shape and texture and 2) be able to be driven by 3D poses. 
Previous works~\cite{mundra2023livehand,chen2023handavatar,corona2022lisa} created a 3D hand avatar from a long capture from a studio~\cite{Moon_2020_ECCV_InterHand2.6M,moon2020deephandmesh} using accurate 3D assets, such as 3D tracking results and calibrated multi-view images.
Assuming such 3D assets is a bottleneck for making a 3D hand avatar in our daily life as capturing and acquiring such 3D assets require lots of resources, such as tens or hundreds of calibrated and synchronized cameras.
Recently, HARP~\cite{karunratanakul2023harp} is introduced, which can make a 3D hand avatar from a short phone scan.
It uses subdivided MANO~\cite{romero2017embodied} as an underlying geometric representation and optimizes albedo and normal maps for personalization.
Compared to HARP, our adaptation pipeline produces more authentic results by utilizing priors from our UHM.

%% file: src/uhm.tex
\section{UHM}

\subsection{Formulation}

\begin{figure}[t]
\begin{center}
\includegraphics[width=0.7\linewidth]{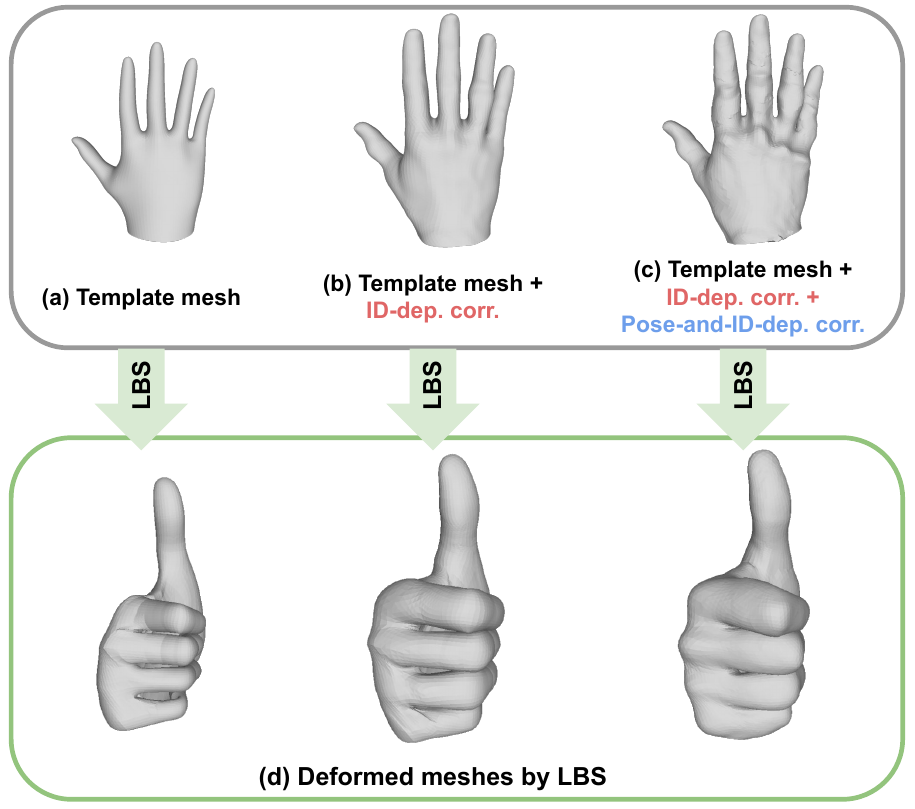}
\end{center}
\vspace*{-6mm}
\caption{
The effectiveness of the correctives.
}
\vspace*{-5mm}
\label{fig:effect_of_corr}
\end{figure}

We use the linear blend skinning (LBS) as an underlying geometric deformation algorithm following previous mesh-based ones~\cite{romero2017embodied,moon2020deephandmesh,li2022nimble,potamias2023handy}.
Given 3D vertices and 3D joint coordinates in the zero pose space (\textit{i.e.}, template space), denoted by $\bar{\bm{J}}$ and $\bar{\bm{V}}$ respectively, we apply various correctives to them and perform LBS to apply the 3D pose to the zero pose space.
Fig.~\ref{fig:effect_of_corr} shows the effects of the correctives.
There are three types of correctives: ID-dependent skeleton corrective $\Delta\bar{\bm{J}}^\text{id}$, ID-dependent vertex corrective $\Delta\bar{\bm{V}}^\text{id}$, and pose-and-ID-dependent vertex corrective $\Delta\bar{\bm{V}}^\text{pose}$.
The ID-dependent skeleton corrective $\Delta\bar{\bm{J}}^\text{id}$ and ID-dependent vertex corrective $\Delta\bar{\bm{V}}^\text{id}$ are to model different 3D skeleton (\textit{e.g.}, bone lengths) and 3D hand shapes (\textit{e.g.}, thickness) in the zero pose space, respectively, for each ID.
The pose-and-ID-dependent vertex corrective $\Delta\bar{\bm{V}}^\text{pose}$ is to model different surface-level deformation mainly driven by 3D poses.
We additionally consider ID to model slightly different pose-dependent vertex corrective for each ID.
To perform LBS, we first perform forward kinematics (FK) with $\bar{\bm{J}}+\Delta\bar{\bm{J}}^\text{id}$ and provided 3D pose $\bm{\Theta}$ to get transformation matrices of each joint.
We denote 3D joint coordinates from FK by $\bm{J}$.
Then, we apply the transformation matrices to $\bar{\bm{V}}+\Delta\bar{\bm{V}}^\text{id}+\Delta\bar{\bm{V}}^\text{pose}$ with pre-defined skinning weights to get final posed 3D mesh $\bm{V}$.
Our template mesh $\bar{\bm{V}}$ consists of 16K vertices and 32K faces.
All three types of correctives are estimated in our pipeline.

\begin{figure}[t]
\begin{center}
\includegraphics[width=\linewidth]{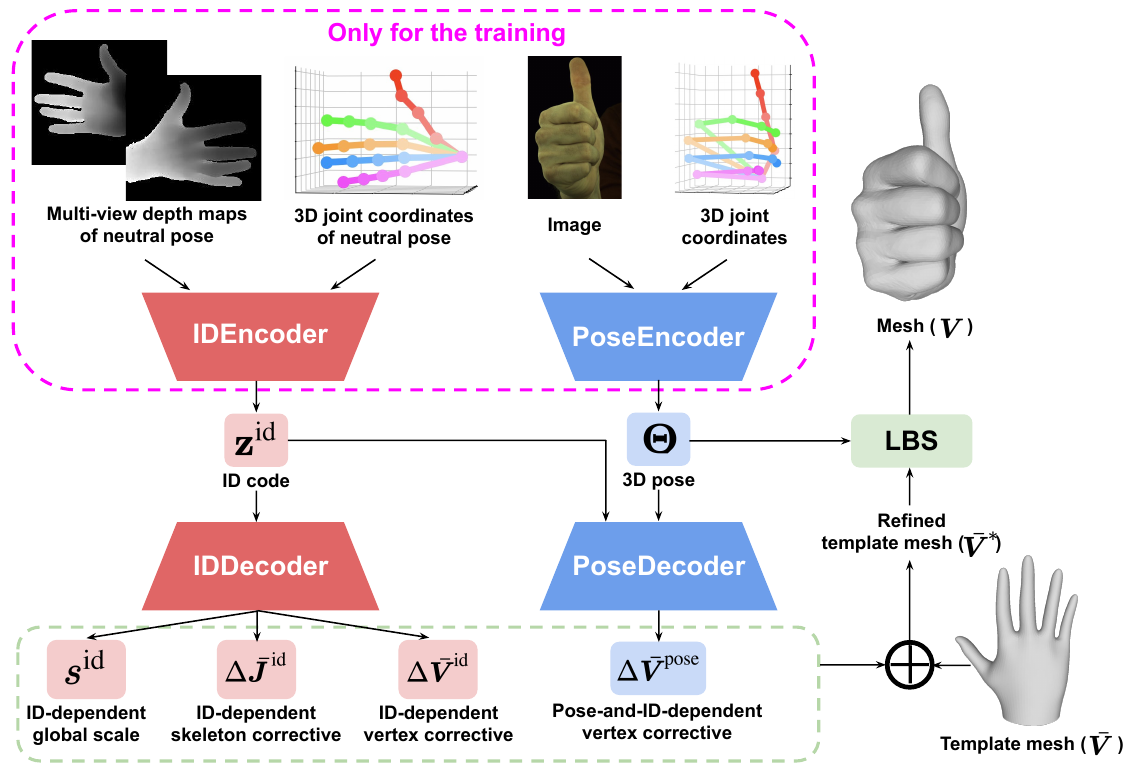}
\end{center}
\vspace*{-7mm}
\caption{
The overall pipeline of the proposed UHM.
The estimated correctives (dotted green box at the bottom) are applied to a template mesh to refine it.
Then, LBS is used to pose the template mesh.
}
\vspace*{-3mm}
\label{fig:overall_pipeline}
\end{figure}

\subsection{Components}\label{subsec:components}
Fig.~\ref{fig:overall_pipeline} shows the overall pipeline of our UHM.
UHM consists of IDEncoder, IDDecoder, PoseEncoder, and PoseDecoder.
Please refer to the supplementary material for their detailed network architectures.

\noindent\textbf{IDEncoder and IDDecoder.}
IDEncoder and IDDecoder are encoder and decoder of variational autoencoder (VAE)~\cite{kingma2014vae}, respectively, responsible for learning priors of the ID space.
IDEncoder extracts ID code $\mathbf{z}^\text{id} \in \mathbb{R}^{32}$ from a pair of a depth map and 3D joint coordinates of each training subject using the reparameterization trick~\cite{kingma2014vae}.
Then, from the ID code, IDDecoder outputs ID-dependent skeleton corrective $\Delta\bar{\bm{J}}^\text{id}$ and ID-dependent vertex correctives $\Delta\bar{\bm{V}}^\text{id}$.
IDEncoder always takes the same inputs for the same subject during the training, and its inputs are prepared by rigidly aligning the 3D scan and 3D joint coordinates of a neutral pose to a reference frame and rendering depth maps from the aligned 3D scan.
In this way, we can normalize pose and viewpoint, not related to the ID information, from the inputs of the IDEncoder.
After the training, the IDEncoder is discarded as inputs of IDEncoder are not affordable for in-the-wild cases.
Instead, we obtain ID codes from novel samples in testing time by fitting ID codes to target data (Sec.~\ref{sec:compare_3d} and ~\ref{sec:compare_adaptation}).

\noindent\textbf{PoseEncoder and PoseDecoder.}
PoseEncoder outputs 6D rotation~\cite{zhou2019continuity} of joints $\bm{\Theta}$ from a pair of a single RGB image and 3D joint coordinates of arbitrary poses and identities.
Unlike IDEncoder's inputs consist of a single pair of each subject, PoseEncoder's inputs are from any poses and subjects.
PoseDecoder outputs pose-and-ID-dependent vertex correctives $\Delta\bar{\bm{V}}^\text{pose}$ from a pair of 6D rotational pose $\bm{\Theta}$ and ID code $\mathbf{z}^\text{id}$ with MLPs.
As how skin deforms can be different for each person even with the same pose, our PoseDecoder takes both pose and ID codes.
Please note that ID-dependent deformations in the zero pose are already covered in IDDecoder, and the role of the additional ID code input to PoseDecoder is to model only different pose-dependent deformations for each ID.
Following STAR~\cite{osman2020star}, we estimate $\Delta\bar{\bm{V}}^\text{pose}$ in a sparse manner with the help of learnable vertex weights $\bm{\Phi}$.
For the same reason as IDEncoder, PoseEncoder is discarded after the training.
In the testing time, we obtain poses from novel samples by fitting them to target data (Sec.~\ref{sec:compare_3d} and ~\ref{sec:compare_adaptation}).

%% file: src/track_and_model.tex
\section{Simultaneous tracking and modeling}\label{sec:track_and_model}

We train UHM in an end-to-end manner from scratch with our simultaneous tracking and modeling pipeline.
There are two types of loss functions that we minimize: data terms and regularizers.
We describe our data terms below and please refer to the supplementary material for the detailed descriptions of the regularizers.

\noindent\textbf{Pose loss, point-to-point loss, and mask loss.}
The pose loss $L_\text{pose}$ is a $L1$ distance between 3D joint coordinates $\bm{J}$ and targets.
It mainly provides information on kinematic deformation.
The point-to-point loss $L_\text{p2p}$ is the closest $L1$ distance 3D vertex coordinates $\bm{V}$ and 3D scans.
The mask loss $L_\text{mask}$ is a $L1$ distance between rendered and target foreground masks, where our masks are from a differentiable renderer~\cite{wei2019vr}.
$L_\text{p2p}$ and $L_\text{mask}$ mainly provides information of non-rigid surface deformation.
For both $L_\text{p2p}$ and $L_\text{mask}$, we calculate the loss functions between two pairs.
First, we use both correctives ($\Delta\bar{\bm{V}}^\text{id}$ and $\Delta\bar{\bm{V}}^\text{pose}$) to obtain $\bm{V}$ and compute the loss functions.
Second, we set $\Delta\bar{\bm{V}}^\text{pose}$ to zero to obtain $\bm{V}$ and compute the loss functions.
The second one enables us to supervise the ID-dependent corrective $\Delta\bar{\bm{V}}^\text{id}$ without being affected by the pose-and-ID-dependent corrective $\Delta\bar{\bm{V}}^\text{pose}$, necessary to learn meaningful ID latent space.

\noindent\textbf{Image matching loss.}
Solely using the above loss functions does not encourage vertices to be semantically consistent across all frames and subjects as both 3D scans and masks are unstructured surface data.
For example, a certain vertex, supposed to be located on the thumbnail across all frames and subjects, could slide to a semantically wrong position.
This is because the above loss functions do not encourage such semantic consistency.
For semantic consistency, we additionally compute an image matching loss, motivated by ~\cite{bogo2014faust,xiang2022dressing}

\begin{figure}[t]
\begin{center}
\includegraphics[width=0.7\linewidth]{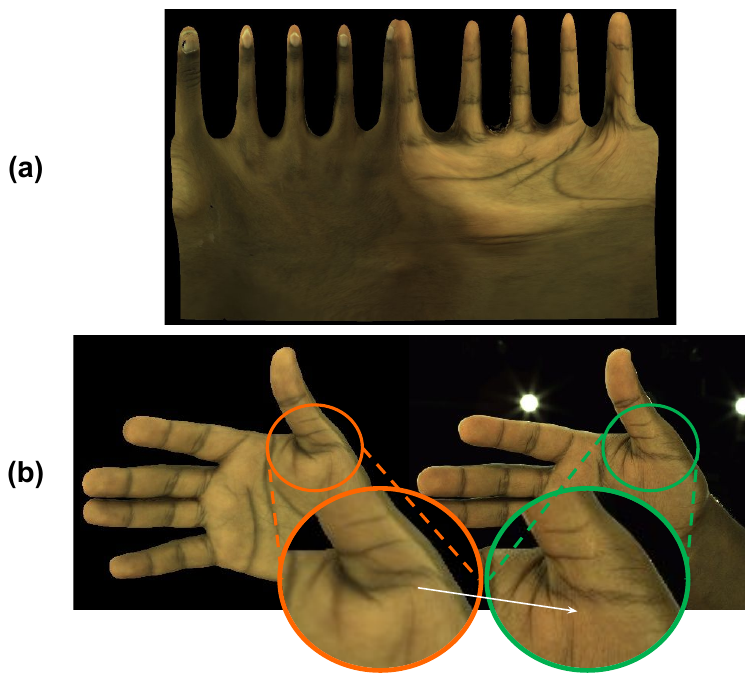}
\end{center}
\vspace*{-7mm}
\caption{
(a) Reference texture.
(b) Our image matching loss function encourages rasterized vertices (orange) to move to the target positions (green), where the target position is obtained by the optical flow (white arrow).
}
\vspace*{-3mm}
\label{fig:image_matching_loss}
\end{figure}

First, for each subject, we unwrap multi-view images of a frame with the neutral pose to UV space, as shown in Fig.~\ref{fig:image_matching_loss} (a), which becomes a \emph{reference texture}.
For the unwrapping, we use our 3D meshes, obtained from a checkpoint that is trained without the image matching loss.
After the unwrapping, we have as many reference textures as there are subjects.
The reference textures are static assets and do not change during the training.
Then, we fine-tune the checkpoint with additional $L_\text{img}$.
Fig.~\ref{fig:image_matching_loss} (b) shows what $L_\text{img}$ does.
We first rasterize mesh vertices and render images~\cite{wei2019vr} using the reference texture (Fig.~\ref{fig:image_matching_loss} (a)) in a differentiable way.
Then, we compute optical flow from the rendered images to captured images using a pre-trained state-of-the-art optical flow estimation network~\cite{teed2020raft}.
Finally, we minimize the $L1$ distance between 1) the 2D positions of the rasterized mesh vertices and 2) the positions of the target pixels, where the target pixels are the output of the optical flow.

Our image matching loss encourages each rasterized mesh vertex to have consistent semantic meanings from that of the reference texture, which results in low variance.
It also results in low bias as the reference texture is from the neutral pose, which has a minimum skin sliding.
Please note that the gradient is only backpropagated to the rasterized mesh vertices.
The rendered images are not perfectly identical to captured images as such rendered images do not have pose-and-view-dependent texture changes and shadow changes.
However, we observed that optical flow is highly robust to such changes in textures, which gives reasonable matching between rendered and captured images.

%% file: src/adaptation_to_a_phone_scan.tex
\section{Adaptation to a phone scan}

After training our UHM following Sec.~\ref{sec:track_and_model}, we adapt it to a short (usually around 15 seconds) phone scan for the authentic hand avatar.
The phone scan includes a single person's hand with the neutral pose and varying global rotations to expose most of the surface of the hand.
During the adaptation, we freeze pre-trained UHM while optimizing its inputs.

\subsection{Preprocessing}\label{sec:adaptation_preprocess}
We use a single iPhone 12 to scan a hand, which incorporates a depth sensor that can be used to extract better geometry of the user’s hand.
Then, we use a 2D hand keypoint detector (our in-house detector or public Mediapipe~\cite{mediapipe}) to obtain 2D hand joint coordinates and RVM~\cite{lin2022robust} to obtain foreground masks.
Also, we use InterWild~\cite{moon2023interwild} to obtain MANO~\cite{romero2017embodied} parameters of all frames.

\begin{figure}[t]
\begin{center}
\includegraphics[width=\linewidth]{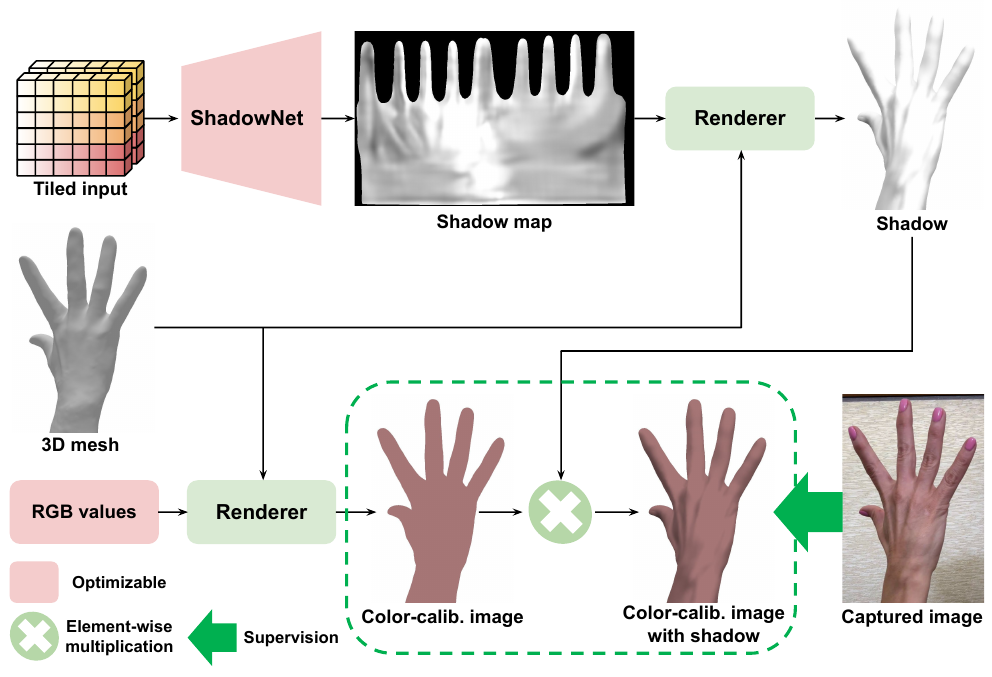}
\end{center}
\vspace*{-7mm}
\caption{
The overall pipeline to remove the shadow from the phone scan using our ShadowNet.
}
\vspace*{-5mm}
\label{fig:shadownet}
\end{figure}

\subsection{Geometry fitting}\label{sec:adaptation_geo_fit}

We fit inputs of our pre-trained UHM (\textit{i.e.}, 3D pose $\bm{\Theta}$ and ID code $\mathbf{z}^\text{id}$), 3D global rotation, and 3D global translation to the phone scan.
The 3D pose, 3D global rotation, and 3D global translation are per-frame parameters, and the ID code is a single parameter and shared across all frames as each phone scan is from a single person.
For the fitting, we minimize loss functions against 2D hand joint coordinates, foreground mask, a depth map, and 3D joint coordinates from the MANO parameters, where the fitting targets are from Sec.~\ref{sec:adaptation_preprocess}.
Please refer to the supplementary material for a detailed description of the fitting.

\subsection{Shadow removal}\label{sec:adaptation_shadow_removal}

\begin{wrapfigure}{r}{0.45\linewidth}
\vspace{-3.5em}
\begin{center}
\includegraphics[width=\linewidth]{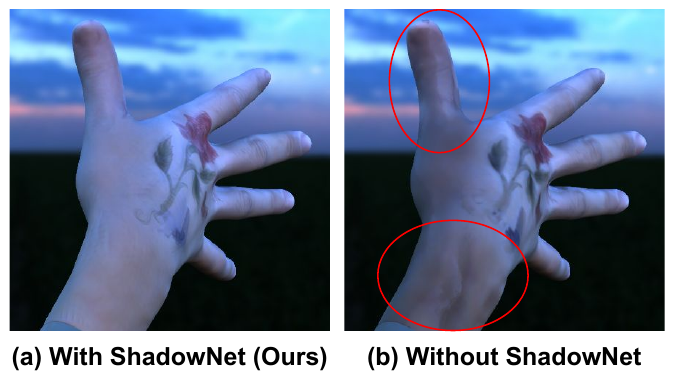}
\end{center}
\vspace{-6mm}
\caption{
Effectiveness of our ShadowNet in a novel light condition.
}
\label{fig:ablation_shadownet}
\end{wrapfigure}
To produce albedo textures, we need to remove shadows from our phone scan.
Fig.~\ref{fig:ablation_shadownet} shows that without removing shadows, the shadow of the phone capture is baked into the texture, which makes significant artifacts in a novel light condition.
Without knowing the full 3D environment map of the phone scan, it is impossible to perfectly disentangle shadow from the unwrapped texture.
Previous work~\cite{karunratanakul2023harp} assumes a single point light and optimizes it during the adaptation.
However, in most cases, the assumption does not hold as there are often more than one light source in our daily life.
Instead of using such a physics-based approach, we use a statistical approach by introducing our ShadowNet.
As shown in Fig.~\ref{fig:shadownet}, our intuition is modeling shadow as a darkness difference between a color-calibrated image and a captured image.

\begin{figure}[t]
\begin{center}
\includegraphics[width=0.9\linewidth]{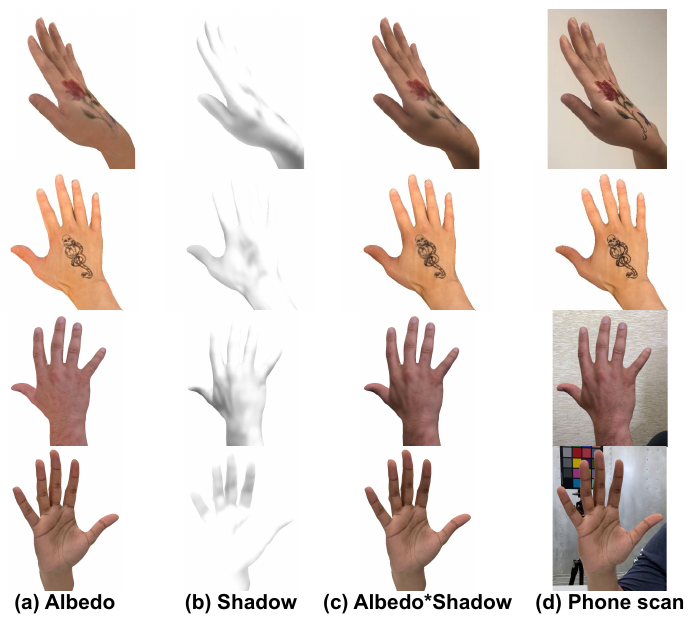}
\end{center}
\vspace*{-7mm}
\caption{
Qualitative results of image's albedo and shadow decomposition using our ShadowNet.
}
\vspace*{-5mm}
\label{fig:albedo_shadow_render}
\end{figure}

\noindent\textbf{ShadowNet.}
Our ShadowNet estimates shadow map in the UV space from tiled 3D global rotation, 3D pose $\bm{\Theta}$, ID code $\mathbf{z}^\text{id}$, and view direction for each mesh vertex.
Given a fixed 3D environment during the phone scan, the inputs of our ShadowNet can determine the shadow of the hand.
The ShadowNet is a fully convolutional network with several upsampling layers.
To encourage smooth shadow, we perform bilinear upsampling four times at the end of the network.
We add a learnable positional encoding to the input before passing it to our ShadowNet as each texel in the UV space has its own semantic meaning.
We apply a sigmoid activation function at the end of our ShadowNet.
By rendering and multiplying our shadow map to an image, we can make the image darker, which can be seen as a shadow casting, similar spirit of Bagautdinov~\etal~\cite{bagautdinov2021driving}.
Fig.~\ref{fig:albedo_shadow_render} shows the qualitative results of our ShadowNet.
We randomly initialize our ShadowNet and train to our phone scan.
Please refer to the supplementary material for the detailed architecture.

\begin{figure}[t]
\begin{center}
\includegraphics[width=0.8\linewidth]{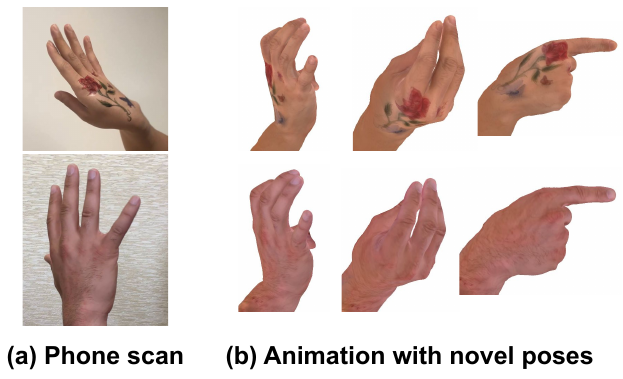}
\end{center}
\vspace*{-7mm}
\caption{
Animated hand avatars whose textures are from (a) phone scan, and geometry is from UHM by passing novel 3D poses $\bm{\Theta}$ and personalized ID code $\mathbf{z}^\text{id}$ to it.
}
\vspace*{-3mm}
\label{fig:animation}
\end{figure}

\begin{table*}[t]
\setlength{\tabcolsep}{1pt}
\begin{minipage}{.6\linewidth}
\centering
\includegraphics[width=\linewidth]{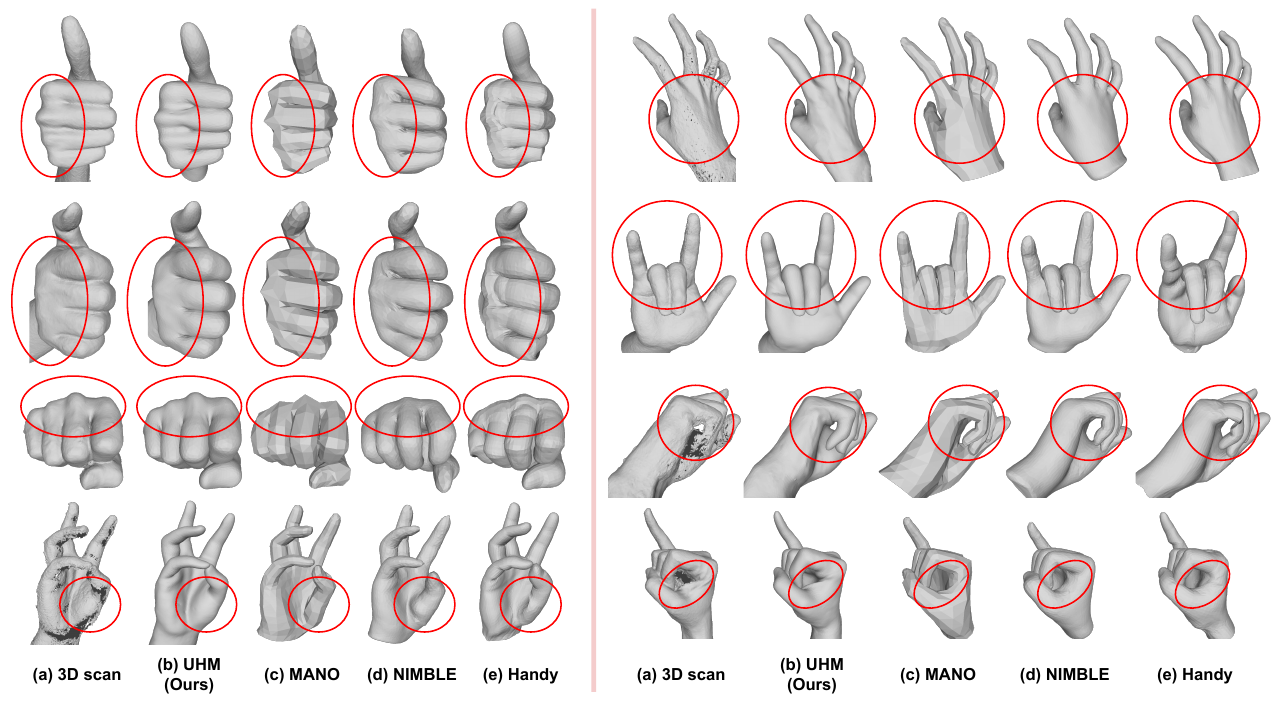}
\vspace*{-8mm}
\captionof{figure}
{
Comparison of our UHM and previous 3D hand models~\cite{romero2017embodied,li2022nimble,potamias2023handy} on our test set.
The first row examples are from the same ID with a sharp hand, and the second row examples are from another same ID with a thick hand.
All the others are from different IDs. 
}
\label{fig:compare_3d}
\end{minipage}\hfill
\vspace*{-3mm}
\begin{minipage}{.37\linewidth}
\centering
\scalebox{0.7}{
\begin{tabular}{C{3.0cm}|C{1.3cm}C{1.3cm}C{1.3cm}}
\specialrule{.1em}{.05em}{.05em}
\multirow{2}{*}{3D hand models} & \multicolumn{3}{c}{Testing sets}  \\
 & Ours & MANO & DHM \\ \hline
MANO~\cite{romero2017embodied} &  1.44 & 0.94 & 1.36 \\
NIMBLE~\cite{li2022nimble} & 1.21 & 0.88 & 1.22 \\
Handy~\cite{potamias2023handy} & 1.20 & 0.78 & 1.11  \\ \hline
\textbf{UHM (low res.)} & 0.73 & 0.76 & 0.61 \\ 
\textbf{UHM (Ours)} & \textbf{0.72} & \textbf{0.75} & \textbf{0.59} \\ 
\specialrule{.1em}{.05em}{.05em}
\end{tabular}
}
\vspace*{-3mm}
\captionof{table}{P2S error (mm) comparison of 3D hand models on multiple test sets.}
\label{table:compare_3d}
\scalebox{0.7}{
\begin{tabular}{C{3.0cm}|C{1.3cm}C{1.3cm}C{1.3cm}}
\specialrule{.1em}{.05em}{.05em}
\multirow{2}{*}{3D hand models} & \multicolumn{3}{c}{\# of views of DHM test set}  \\
 & 1 view & 2 views & 4 views \\ \hline
LISA~\cite{corona2022lisa} &  3.68 & 3.56 & 3.38 \\
\textbf{UHM (Ours)} & \textbf{1.63} & \textbf{1.38} & \textbf{1.27} \\ 
\specialrule{.1em}{.05em}{.05em}
\end{tabular}
}
\vspace*{-3mm}
\captionof{table}{P2S error (mm) comparison on DHM dataset.}
\label{table:compare_lisa_3d}
\end{minipage}
\end{table*}

\noindent\textbf{Optimization.}
First, we obtain the color-calibrated image, rendered from a UV texture that has the same color for all texels.
The RGB values (3D vector) of texels are optimizable.
Our assumption for the shadow removal is that hands mostly have uniform skin color, unlike the human body with different colors in upper and lower body clothes.
Please note that we use the color-calibrated image only for removing shadow, and our final hand avatar has authentic information from any colors.

Then, we multiply the rendered shadow to the color-calibrated image.
We minimize $L1$ distance and VGG loss~\cite{ledig2017photo} between two pairs at the same time: between 1) color-calibrated image and captured image and 2) color-calibrated image with shadow and captured image.
In this way, we can optimize ShadowNet to produce the 1-channel difference between the captured image and color-calibrated image following the image intrinsic decomposition formula.
Without proper regularizers, our ShadowNet can consider all 1-channel differences as a shadow, which is not desirable for hair and black tattoos.
Hence, we apply a total variation regularizer to the rendered shadow to model shadow as a \emph{locally smooth} darkness changes without locally sharp ones.

\subsection{Texture optimization}\label{sec:adaptation_texture_optimization}

Given estimated 3D meshes from Sec.~\ref{sec:adaptation_geo_fit} and shadow from Sec.~\ref{sec:adaptation_shadow_removal}, we first divide captured images by the shadow and unwrap them to UV space.
Then, we average them considering the visibility of each texel.
We preprocess the unwrapped texture with the OpenCV inpainting function to fill missed texels.
To further optimize the unwrapped texture, we render an image from the unwrapped texture and multiply the rendered shadow to it.
Then, we minimize $L1$ distance and VGG loss~\cite{ledig2017photo} between the rendered image and captured images for a more photorealistic texture.
We additionally encourage locally smooth textures for missing texels, inpainted by OpenCV.
During the texture optimization, we fine-tune our ShadowNet to make the shadow consistent with our texture.

\subsection{Final outputs}
The final outputs of our hand avatar creation pipeline are 1) optimized ID code of UHM $\mathbf{z}^\text{id}$ from Sec.~\ref{sec:adaptation_geo_fit} and 2) optimized texture from Sec.~\ref{sec:adaptation_texture_optimization}.
The geometry ID code gives a personalized 3D hand shape and skeleton, and the optimized texture provides personalized albedo texture.
By feeding 3D poses from off-the-shelf 3D hand pose estimators~\cite{ge20193d,choi2020p2m,moon2020i2l,lin2021end,lin2021mesh,moon2022hand4whole} with the optimized ID code to pre-trained UHM, entire mesh vertices can be animated from the novel poses.
Also, simply using the standard computer graphics pipeline, authentic 3D hand avatars can be rendered with the personalized albedo texture, as shown in Fig.~\ref{fig:animation}, or with Phong reflection model, as shown in Fig.~\ref{fig:teaser} (b).
Our pipeline takes 2 hours for 15 seconds of phone scan, while HARP takes 6 hours.

%% file: src/experiment.tex
\begin{table*}[t]
\setlength{\tabcolsep}{1pt}
\begin{minipage}{.52\linewidth}
\centering
\includegraphics[width=\linewidth]{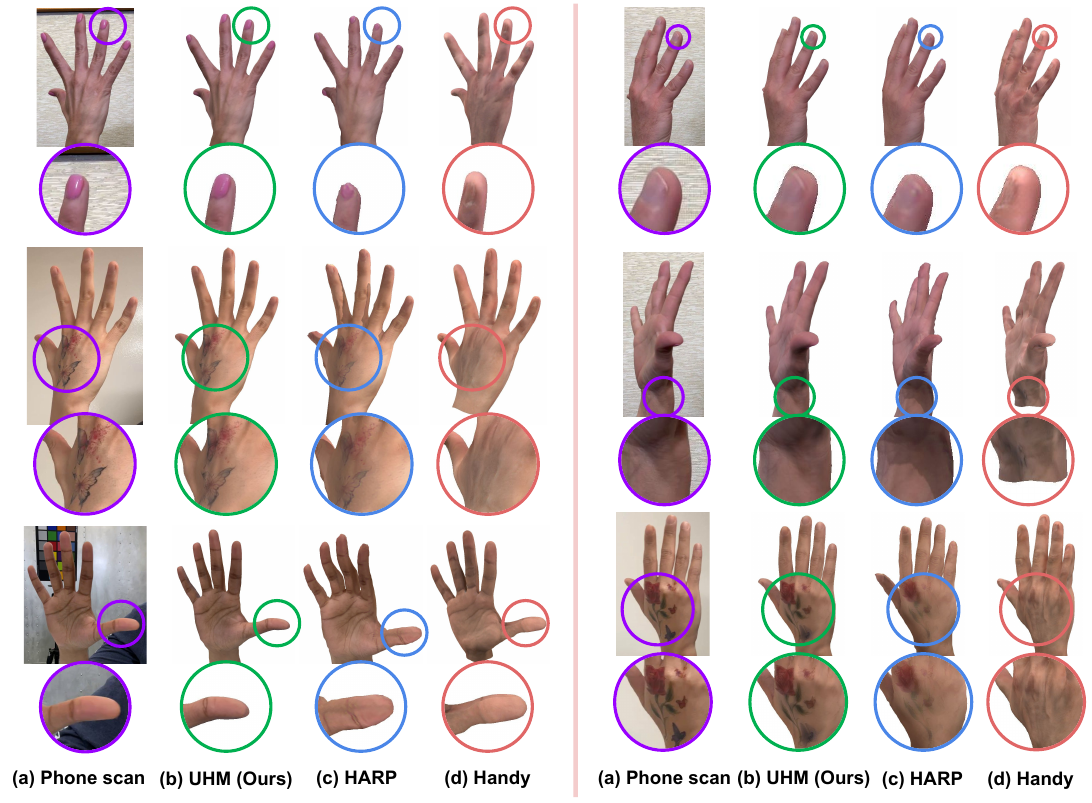}
\vspace*{-8mm}
\captionof{figure}
{
Comparison of various hand avatars on the training set of our phone scan dataset.
}
\vspace*{-3mm}
\label{fig:compare_adaptation}
\end{minipage}\hfill
\vspace*{-3mm}
\begin{minipage}{.47\linewidth}
\centering
\includegraphics[width=\linewidth]{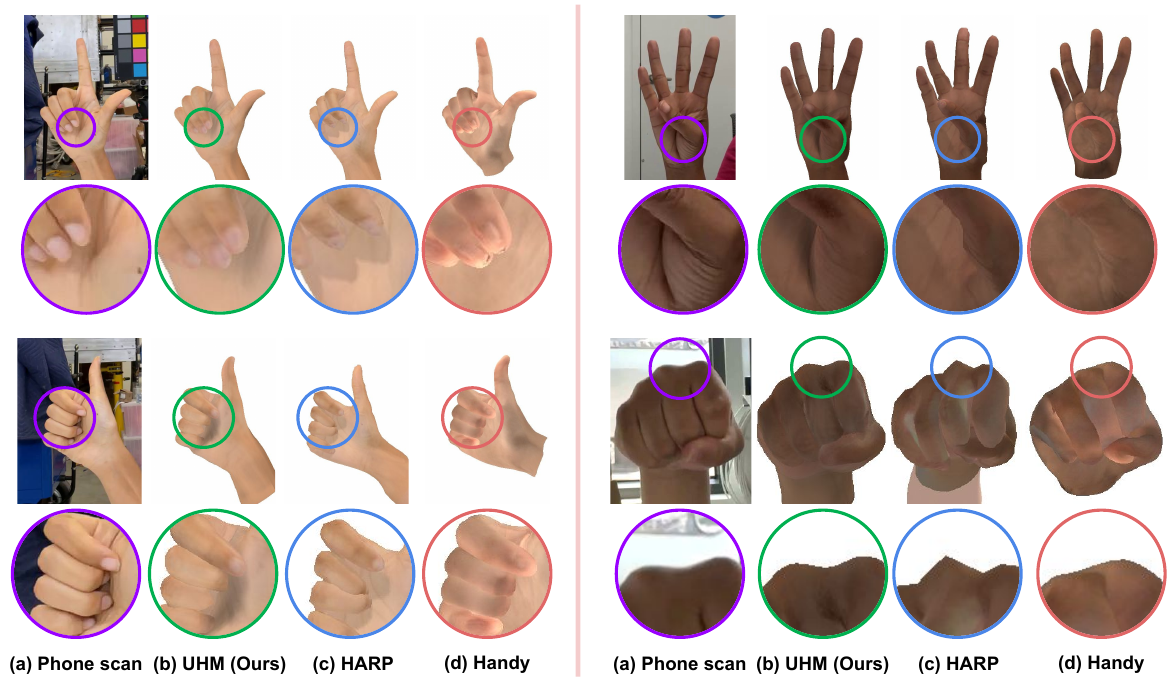}
\vspace*{-8mm}
\captionof{figure}
{
Comparison of various hand avatars on the testing set of our phone scan dataset.
}
\label{fig:compare_adaptation_test}
\scalebox{0.7}{
\begin{tabular}{C{2.8cm}|C{1.3cm}C{1.3cm}C{1.3cm}C{1.3cm}}
\specialrule{.1em}{.05em}{.05em}
3D hand avatars &  PSNR\textuparrow & SSIM\textuparrow & LPIPS\textdownarrow & P2S\textdownarrow \\ \hline
Handy~\cite{potamias2023handy} & 26.02 & 0.930 & 0.134 &  2.21 \\ 
HARP~\cite{karunratanakul2023harp} & 29.89 & 0.952 & 0.092 & 2.04 \\ 
\textbf{UHM (Ours)} & \textbf{31.82} & \textbf{0.962} & \textbf{0.076} & \textbf{0.45} \\ 
\specialrule{.1em}{.05em}{.05em}
\end{tabular}
}
\vspace*{-3mm}
\captionof{table}{Comparison of 3D hand avatars on our test set.}
\vspace*{-3mm}
\label{table:compare_adaptation_our_data}
\end{minipage}
\end{table*}

\begin{table}[t]
\scriptsize
\centering
\setlength\tabcolsep{1.0pt}
\def\arraystretch{1.1}
\begin{tabular}{C{2.4cm}|C{1.2cm}C{1.2cm}C{1.3cm}}
\specialrule{.1em}{.05em}{.05em}
3D hand avatars &  PSNR\textuparrow & SSIM\textuparrow & LPIPS\textdownarrow \\ \hline
Handy~\cite{potamias2023handy} & 26.10  & 0.930 & 0.087 \\ 
HARP~\cite{karunratanakul2023harp} & 27.50 & 0.947 & 0.081 \\ 
\textbf{UHM (Ours)} & \textbf{32.55} & \textbf{0.957} & \textbf{0.055} \\ 
\specialrule{.1em}{.05em}{.05em}
\end{tabular}
\vspace*{-3mm}
\caption{Comparison of 3D hand avatars on the test set of HARP dataset.}
\vspace*{-5mm}
\label{table:compare_adaptation_harp_data}
\end{table}

\section{Experiments}

\subsection{Datasets}

We use the three datasets below to train and evaluate our UHM.

\noindent\textbf{Our studio dataset.}
We use 177 captures for the training and 7 captures for the testing, where each capture includes 18K frames of a unique subject taken from 170 cameras on average.
The testing subjects are not included in the training set.
Please refer to the supplementary material for the detailed descriptions of our dataset.

\noindent\textbf{Testing set of MANO.}
We report 3D errors on the testing set of MANO, which consists of 50 3D scans from 6 subjects.
It is used only for the evaluation purpose.

\noindent\textbf{Dataset of DHM.}
We report 3D errors on the dataset of DHM, which consists of 33K 3D scans from a single subject.
We use this dataset only for the evaluation purpose.

We also use the two datasets below to evaluate the adaptation pipeline.

\noindent\textbf{Our new phone scan dataset.}
We newly captured 18 phone scans from unique IDs and use them to evaluate our adaptation pipeline.
We use 4 scans out of 18 scans for the quantitative evaluations.
For the training, frames with neutral poses are used, and for the testing, frames with diverse poses are used.
All the phone scans are preprocessed following Sec.~\ref{sec:adaptation_preprocess}.
Some phone scans have distinctive authenticities, such as fingernail polish and tattoos.
Please refer to the supplementary material for the detailed descriptions of our dataset.

\noindent\textbf{Dataset of HARP.}
We report errors in the publicly available HARP dataset.
Please note that they only released a partial of what they used in paper, and the released one consists of a single ID.
For the quantitative results, we used \textit{subject\_1} sequence as all other sequences do not have enough pose diversity, which cannot be used for the testing.
Among 9 sub-sequences of \textit{subject\_1}, 1 to 5 are used for the training, and 6 to 9 are used for the testing.

\subsection{Comparison of 3D hand models}\label{sec:compare_3d}
We compare the generalizability of pre-trained 3D hand models to unseen IDs and poses.
To this end, we fit inputs of 3D hand models (\textit{i.e.}, pose and ID code) to target data while fixing the pre-trained 3D hand models.
After fitting them to target data, we measure point-to-surface (P2S) error (mm), which measures the average distance from points of the 3D scan to the surfaces of the output meshes.
The errors are measured after fitting inputs of 3D hand models as much as possible to target data while fixing the models.
In this way, we can check how much fidelity (\textit{i.e.}, surface expressiveness) of each hand model is not enough to fully replicate 3D scans after marginalizing fitting errors.
For UHM, we excluded vertices on the forearm when calculating the error as all others do not have the forearm.
We do not include personalized 3D hand models~\cite{moon2020deephandmesh,mundra2023livehand,chen2023handavatar,iwase2023relightablehands} in the comparisons as our focus in this experiment is to compare generalizability to unseen poses and IDs, while such personalized models cannot generalize to novel IDs.

Fig.~\ref{fig:compare_3d} and Table~\ref{table:compare_3d} show that our UHM produces the best quality of meshes on multiple test sets than other universal hand models, such as MANO~\cite{romero2017embodied}, NIMBLE~\cite{li2022nimble}, and Handy~\cite{potamias2023handy}.
Handy~\cite{potamias2023handy} suffers from surface artifacts.
For example, there are severe artifacts around the knuckle area in the examples at the top three rows and the first column.
Also, there is no muscle bulging around the thumb in the example at the bottom and the first column.
There is a severe artifact at the pinky finger in the example in the third row and the second column.
We additionally provide our results from a low-resolution template, which has half the number of vertices (3K) than NIMBLE (6K) and Handy (7K) for a more fair comparison.
The table demonstrates that even with a half number of vertices, ours achieves better fidelity than NIMBLE and Handy.
Table~\ref{table:compare_lisa_3d} shows that ours achieves much better results on the DHM dataset than LISA~\cite{corona2022lisa}.

\subsection{Comparison of adaptation pipelines}\label{sec:compare_adaptation}

Fig.~\ref{fig:compare_adaptation} and ~\ref{fig:compare_adaptation_test} show that our adaptation pipeline achieves much more authentic and photorealistic results than HARP~\cite{karunratanakul2023harp} and Handy~\cite{potamias2023handy}.
In particular, the right column of Fig.~\ref{fig:compare_adaptation_test} shows that only our avatar has skin bulging around the thumb and sharp knuckle, unseen during the training, thanks to our high-fidelity UHM.
HARP suffers from geometry artifacts, which result in texture artifacts.
We think this is because of the limited expressiveness of the MANO model.
In addition, due to their single point light assumption, they have a clearly different shadow from the captured images, as the second row examples of Fig.~\ref{fig:compare_adaptation} show.
We address such a failure case by introducing the ShadowNet.
Handy suffers from a lack of texture authenticity, such as different fingernail polish, tattoos, and palm wrinkles, as their textures are from pre-defined texture space.
On the other hand, we unwrap textures and directly optimize them without being constrained in texture space, which gives authentic textures.
Unlike geometry, there can be numerous variants in the texture space including shadow, tattoo, and fingernail polish; hence, we think such texture prior is not enough for the authenticity.

\begin{figure}[t]
\begin{center}
\includegraphics[width=0.9\linewidth]{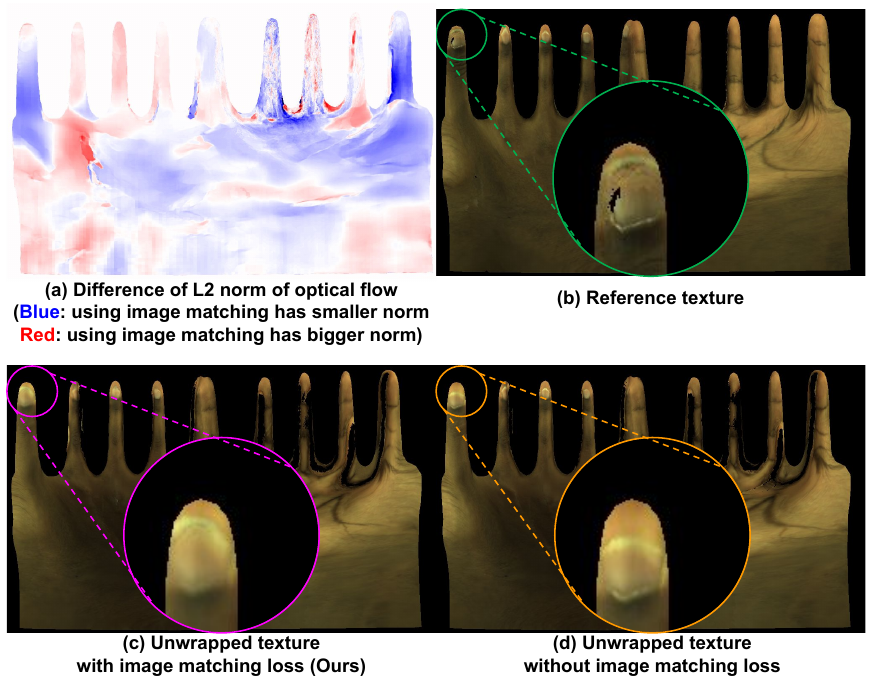}
\end{center}
\vspace*{-7mm}
\caption{
Effectiveness of our image matching loss function.
}
\vspace*{-3mm}
\label{fig:image_matching_ablation}
\end{figure}

Tab.~\ref{table:compare_adaptation_our_data} and ~\ref{table:compare_adaptation_harp_data} show that our adaptation pipeline achieves better numbers.
For a fair comparison, all avatars in Tab.~\ref{table:compare_adaptation_our_data} are trained with the additional depth map loss as our dataset provides depth maps.
For four subjects in our phone scan, we co-captured studio data, which gives 3D data of them.
To measure the accuracy of the adaptation pipeline more thoroughly, we measure the P2S error (mm) between personalized meshes from the phone scan and the 3D scan from our capture studio.
Thanks to our high-fidelity universal modeling, the proposed UHM clearly achieves the best result in the 3D metric.

For the results on the testing set, following the previous protocols~\cite{karunratanakul2023harp} that optimizes 3D poses of hands, lights, and ambient ratio on the testing set, we fine-tune PoseNet and ShadowNet on the test set.
All remaining parameters, including the ID code and optimized texture, are fixed in the testing stage following HARP~\cite{karunratanakul2023harp}.
For the results of HARP, we used their official code with groundtruth hand boxes.
For the results of Handy, we downloaded their official pre-trained weights and optimized 3D pose and texture latent code using $L1$ distance and LPIPS~\cite{zhang2018unreasonable} following their paper.
Please refer to the supplementary material for the detailed fitting process of Handy.

\begin{figure}[t]
\begin{center}
\includegraphics[width=0.8\linewidth]{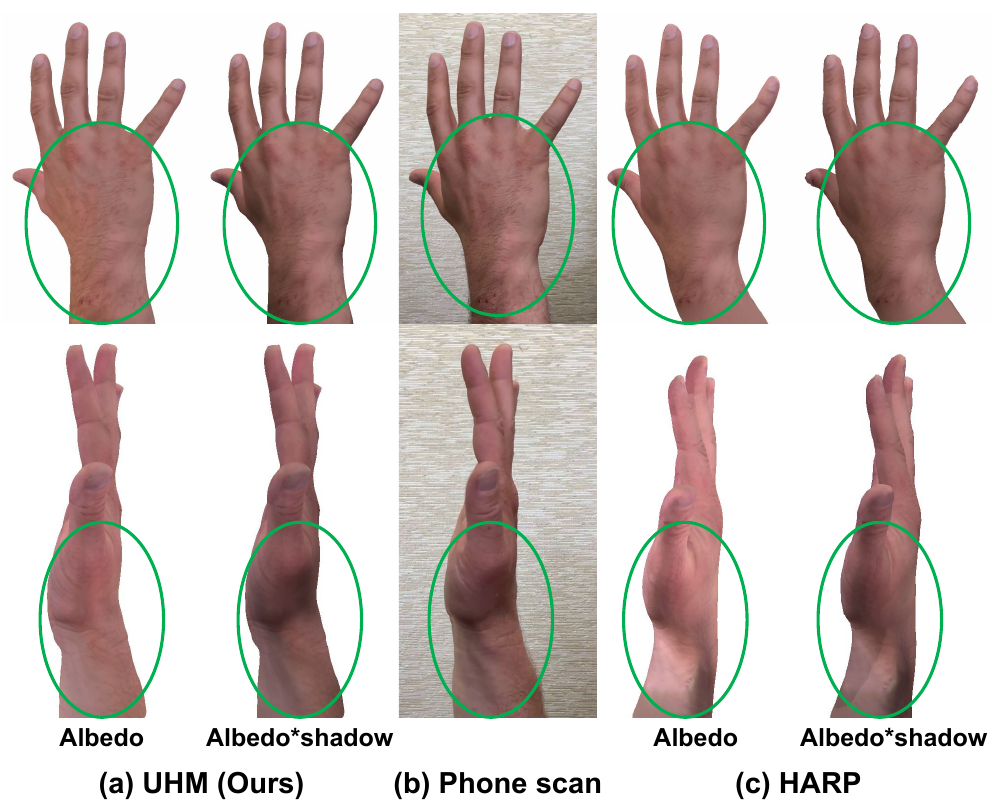}
\end{center}
\vspace*{-7mm}
\caption{
Comparison of rendered images 1) only using albedo and 2) using both albedo and shadow.
}
\vspace*{-3mm}
\label{fig:compare_albedo}
\end{figure}

\subsection{Ablation study}

\noindent\textbf{Image matching loss.}
To validate the effectiveness of our image matching loss $L_\text{img}$ during the tracking and modeling, depicted in Fig.~\ref{fig:image_matching_loss}, we first unwrap multi-view images to UV space using our 3D meshes.
Then, we compute optical flow~\cite{teed2020raft} from the reference texture of the neutral pose (Fig.~\ref{fig:image_matching_ablation} (b)) to the unwrapped per-frame texture.
Fig.~\ref{fig:image_matching_ablation} (a) shows that using our image matching loss $L_\text{img}$ decreases the $L2$ norm of the optical flow for most texels, which shows that texels are located in semantically correct and consistent positions by suffering less from the skin sliding.
In particular, texels that have semantically distinctive locations, such as wrinkles on the palm and thumbnail, have significantly less $L2$ norm of the optical flow as the optical flow provides meaningful correspondences for such texels.
Fig.~\ref{fig:image_matching_ablation} (c) and (d) show that compared to Fig.~\ref{fig:image_matching_ablation} (b), using our image matching loss produces consistent and correct position of thumb in the UV space.
On the other hand, as the back of the hand usually does not have distinctive textures, optical flow fails to produce meaningful correspondence, which results in a slightly higher $L2$ norm.

\noindent\textbf{ShadowNet.}
Fig.~\ref{fig:compare_albedo} shows that the albedo rendering of HARP still has a shadow, while ours does not.
This shows the benefit of using our ShadowNet to remove the shadow from phone scans instead of assuming a single point light and optimizing it like HARP.
In addition, our albedo has more detailed textures, such as hair on the back of the hand (first row).
Due to the ambiguity of the image's intrinsic decomposition, we could not include quantitative evaluations.

%% file: src/conclusion.tex
\section{Conclusion}

We present UHM, a universal hand model that 1) can represent high-fidelity 3D hand mesh of arbitrary IDs and diverse poses and 2) can be adapted to each person with a short phone scan for the authentic 3D hand avatar.
UHM performs the tracking and modeling at the same time to address the error accumulation problem from the tracking stage.
In addition, we newly introduce the image matching loss function to prevent skin sliding during the tracking and modeling.
Finally, our adaptation pipeline achieves a highly authentic hand avatar by utilizing useful learned priors of UHM.

%% file: src_suppl/main_suppl.tex
\begin{center}
\textbf{\large Supplementary Material for \\ ``Authentic Hand Avatar from a Phone Scan via Universal Hand Model"}
\end{center}

\setcounter{section}{0}
\setcounter{table}{0}
\setcounter{figure}{0}

\renewcommand{\thesection}{\Alph{section}}   
\renewcommand{\thetable}{\Alph{table}}   
\renewcommand{\thefigure}{\Alph{figure}}

In this supplementary material, we provide more experiments, discussions, and other details that could not be included in the main text due to the lack of pages.
The contents are summarized below:
\begin{enumerate}[nosep, label=\Alph*.]  
    \item Sec.~\ref{sec:qualitative_results_suppl}: More qualitative results
    \item Sec.~\ref{sec:ablation_study_suppl}: More ablation studies
    \item Sec.~\ref{sec:network_architecture_and_regularizers_suppl}: UHM architectures and loss functions
    \item Sec.~\ref{sec:adaptation_suppl}: Details of adaptation to a phone scan
    \item Sec.~\ref{sec:our_datasets_suppl}: Our datasets
    \item Sec.~\ref{sec:experiment_details_suppl}: Experiment details
    \item Sec.~\ref{sec:failure_cases_suppl}: Failure cases
\end{enumerate}

%% file: src_suppl/more_qualitative_results.tex
\begin{figure*}[t]
\begin{center}
\includegraphics[width=\linewidth]{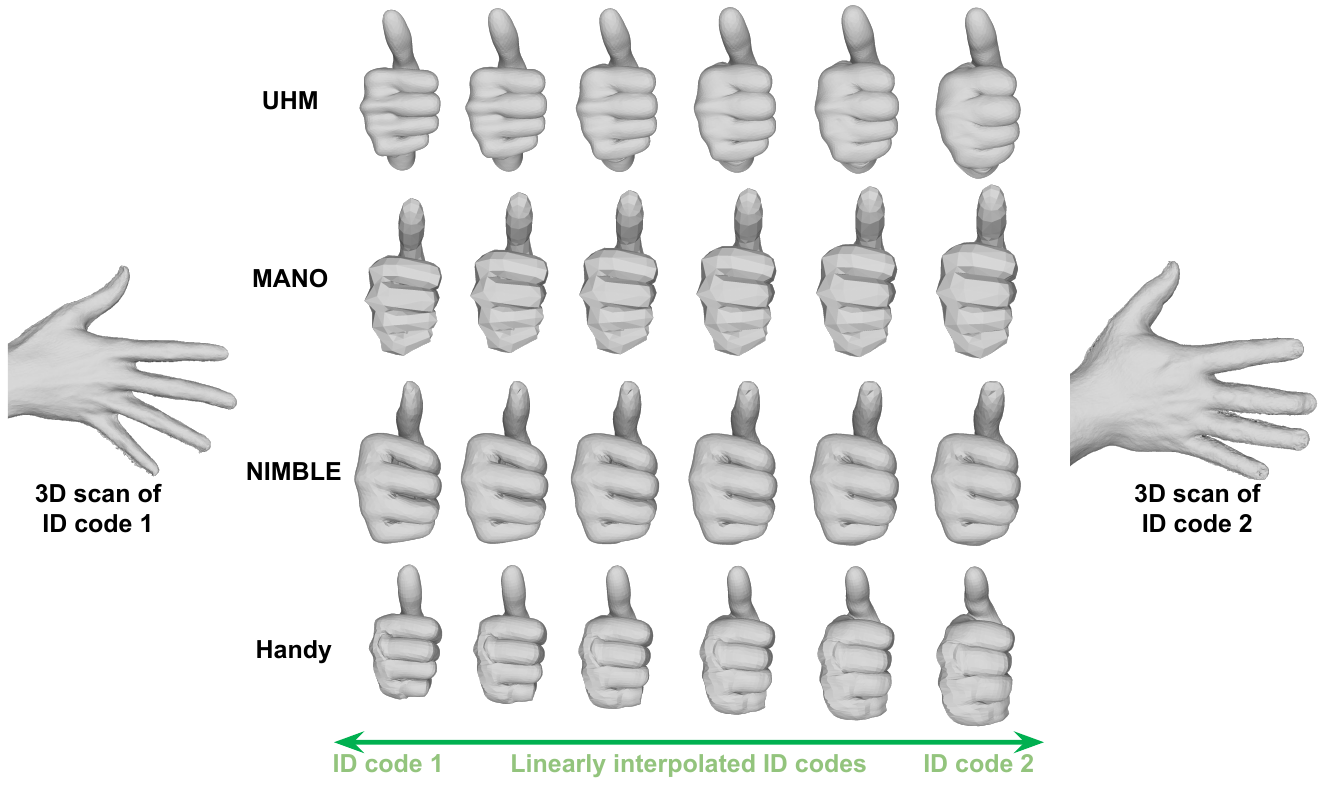}
\end{center}
\vspace*{-7mm}
\caption{
Comparison of 3D meshes from linearly interpolated ID codes.
The leftmost and rightmost 3D scans show examples of the ID codes 1 and 2.
For each row, 3D meshes have the same 3D pose and only ID code changes by a linear interpolation.
}
\vspace*{-3mm}
\label{fig:compare_id_interpolation_1}
\end{figure*}

\begin{figure*}[t]
\begin{center}
\includegraphics[width=\linewidth]{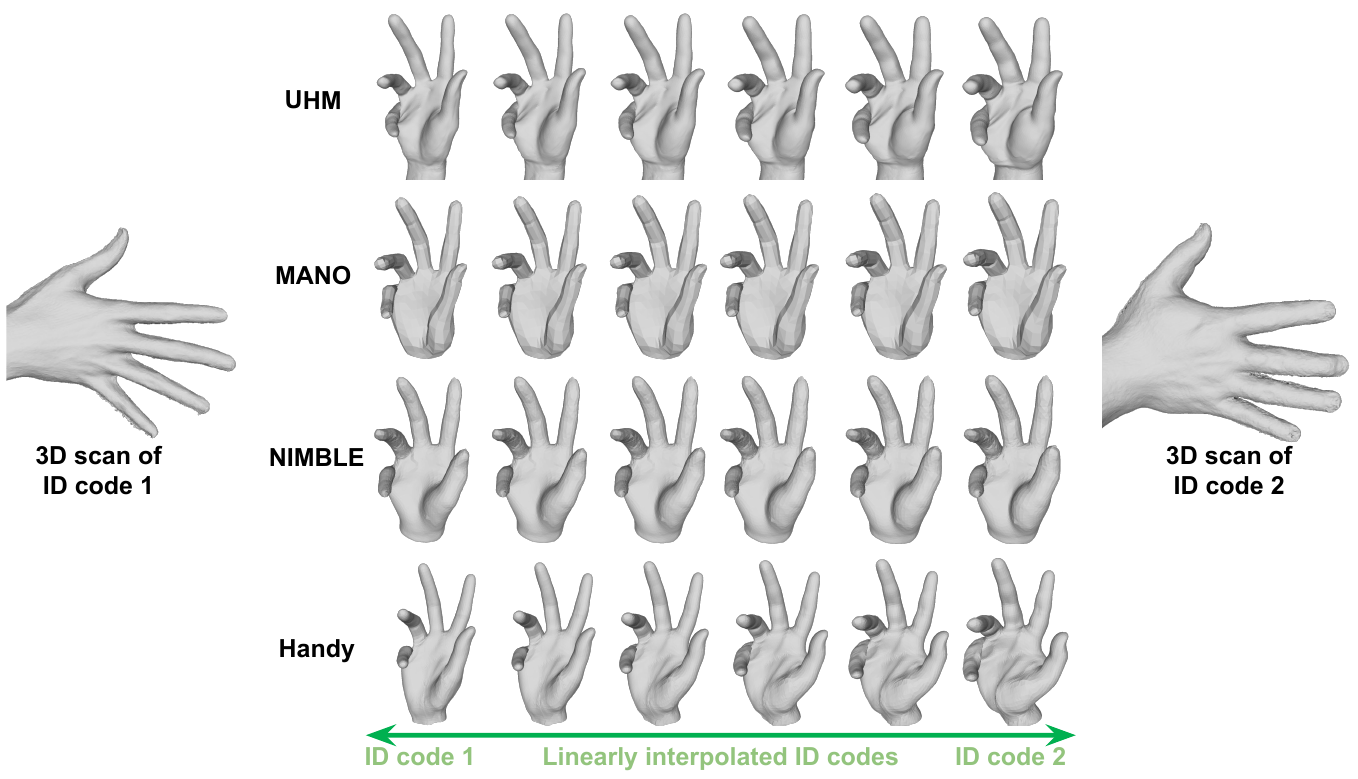}
\end{center}
\vspace*{-7mm}
\caption{
Comparison of 3D meshes from linearly interpolated ID codes.
The leftmost and rightmost 3D scans show examples of the ID codes 1 and 2.
For each row, 3D meshes have the same 3D pose and only ID code changes by a linear interpolation.
}
\vspace*{-3mm}
\label{fig:compare_id_interpolation_2}
\end{figure*}

\begin{figure*}[t]
\begin{center}
\includegraphics[width=\linewidth]{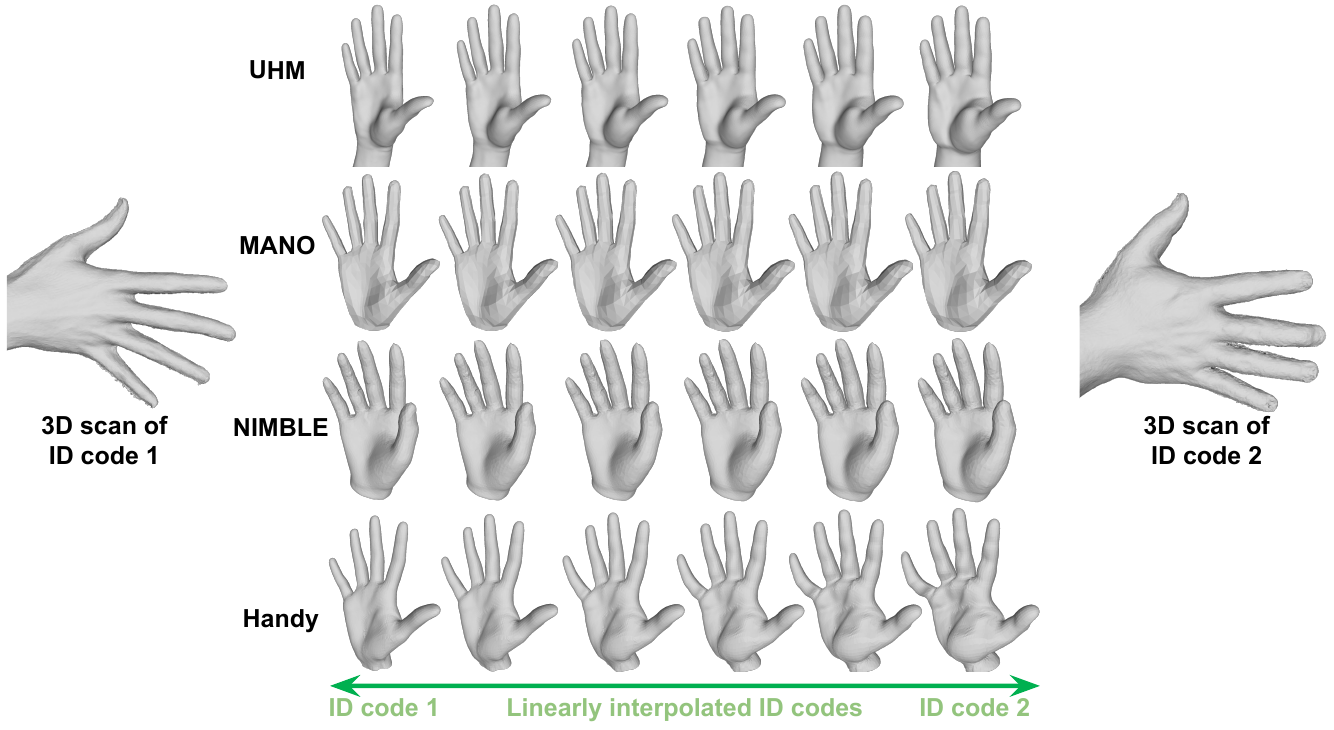}
\end{center}
\vspace*{-7mm}
\caption{
Comparison of 3D meshes from linearly interpolated ID codes.
The leftmost and rightmost 3D scans show examples of the ID codes 1 and 2.
For each row, 3D meshes have the \textbf{zero pose (\textit{i.e.}, 3D pose of the template space)}, and only ID code changes by a linear interpolation.
}
\vspace*{-3mm}
\label{fig:compare_id_interpolation_3}
\end{figure*}

\section{More qualitative results}\label{sec:qualitative_results_suppl}

\subsection{ID code interpolation}
Fig.~\ref{fig:compare_id_interpolation_1}, ~\ref{fig:compare_id_interpolation_2}, and ~\ref{fig:compare_id_interpolation_3} show that our UHM produces smoothly changing 3D meshes from the linearly interpolated ID codes, where the two ID codes are from the unseen test set.
Our 3D meshes from the interpolated ID codes have a natural and realistic surface.
On the other hand, Fig.~\ref{fig:compare_id_interpolation_2} and ~\ref{fig:compare_id_interpolation_3} show that Handy~\cite{potamias2023handy} fails to disentangle 3D pose and ID.
As each row of all figures is from the same pose but from different ID codes, only ID-related information (\textit{i.e.}, thickness) should change while preserving the 3D pose.
However, Fig.~\ref{fig:compare_id_interpolation_2} and ~\ref{fig:compare_id_interpolation_3} show that only changing the ID code of Handy produces 3D meshes with different 3D poses.
This is evident in Fig.~\ref{fig:compare_id_interpolation_3} as the rightmost result of Handy has a totally different 3D pose from the leftmost one.

\begin{figure*}[t]
\begin{center}
\includegraphics[width=\linewidth]{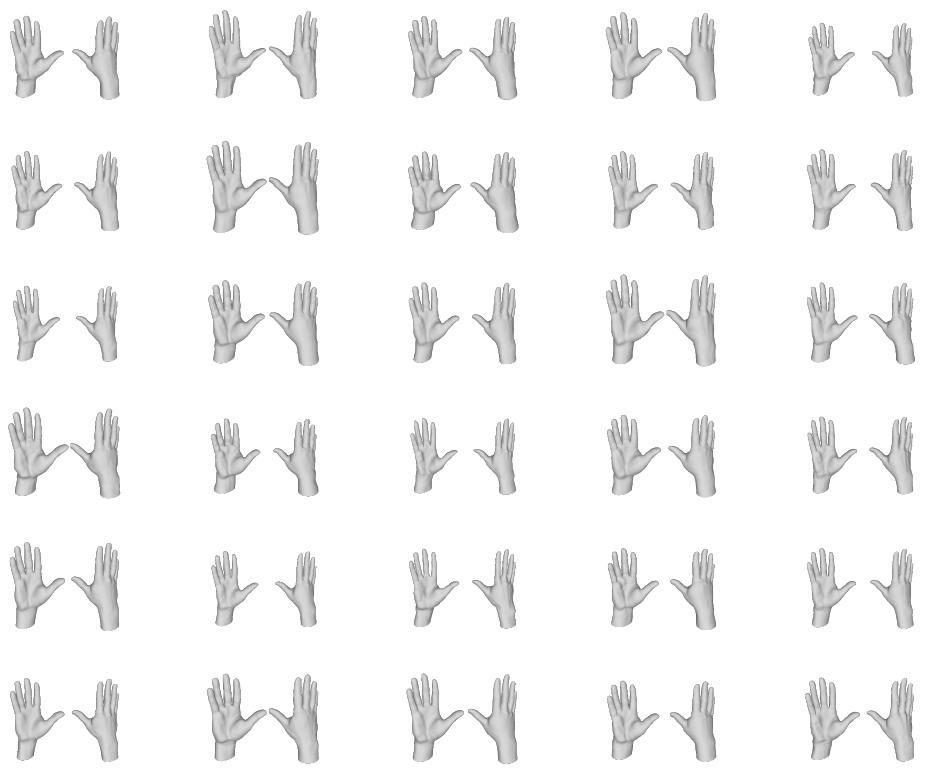}
\end{center}
\vspace*{-7mm}
\caption{
3D meshes from randomly sampled ID codes from the Gaussian distribution with zero 3D poses.
}
\vspace*{-3mm}
\label{fig:id_random_sample}
\end{figure*}

\subsection{ID code random sampling}
Fig.~\ref{fig:id_random_sample} shows 3D meshes from our UHM with randomly sampled ID codes from the Gaussian distribution and zero 3D poses.
Our ID space spans a wide range of ID space, including diverse bone lengths and 3D hand shapes.

\begin{figure*}[t]
\begin{center}
\includegraphics[width=\linewidth]{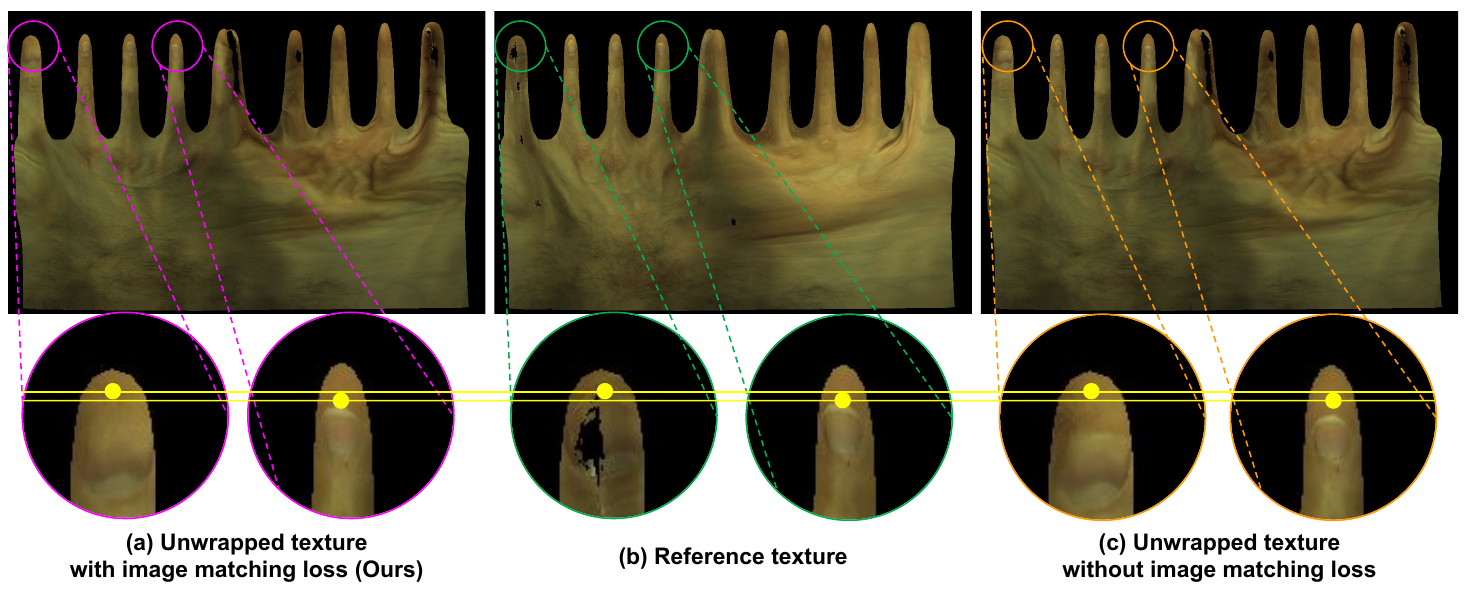}
\end{center}
\vspace*{-7mm}
\caption{
Comparison of 3D hand avatars on HARP dataset~\cite{karunratanakul2023harp}.
}
\vspace*{-3mm}
\label{fig:qualitative_image_matching_loss}
\end{figure*}

\subsection{Effectiveness of the image matching loss}
Fig.~\ref{fig:qualitative_image_matching_loss} shows that using our image matching loss of Sec.~\textcolor{red}{4} of the main manuscript produces consistent unwrapped textures compared to the reference texture.
(a) has consistent fingernail tips (yellow circles), while (c) produces inconsistent ones compared to those of the (b) reference texture.

\begin{figure}[t]
\begin{center}
\includegraphics[width=\linewidth]{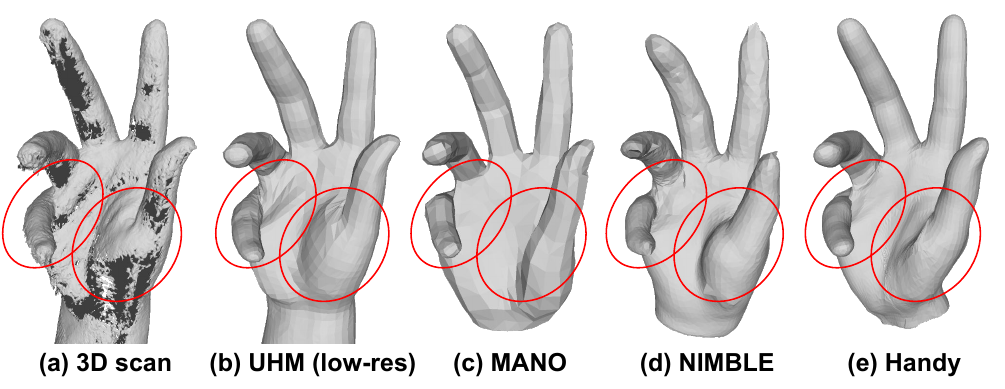}
\end{center}
\vspace*{-3mm}
\captionof{figure}
{
Comparison of the low-resolution UHM and previous 3D hand models ~\cite{romero2017embodied,li2022nimble,potamias2023handy} on our test set.
}
\label{fig:compare_3d_lowres}
\end{figure}

\subsection{Low-resolution UHM}
Fig.~\ref{fig:compare_3d_lowres} demonstrate that even with a half number of vertices (3K), ours achieves better fidelity than NIMBLE (6K) and Handy (7K).
For example, the low-resolution UHM has natural muscle bulging around the thumb and wrinkles around the pinky finger.

\begin{figure*}[t]
\begin{center}
\includegraphics[width=\linewidth]{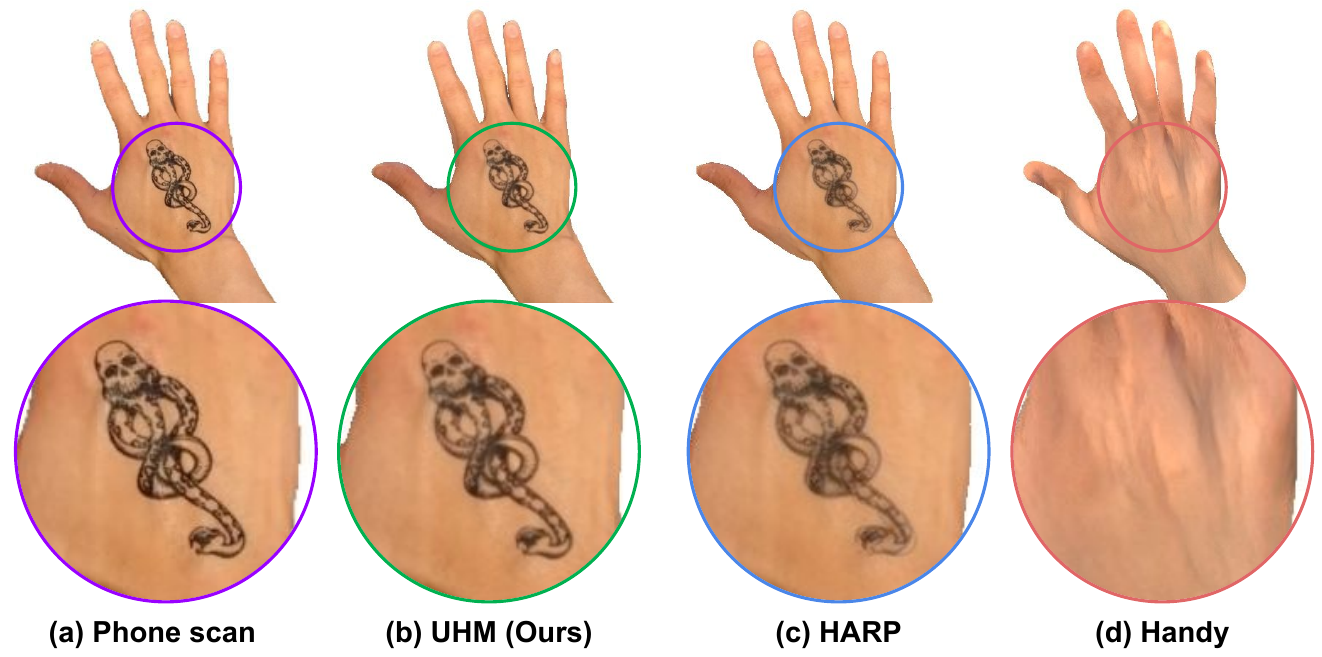}
\end{center}
\vspace*{-7mm}
\caption{
Comparison of 3D hand avatars on HARP dataset~\cite{karunratanakul2023harp}.
}
\vspace*{-3mm}
\label{fig:compare_adaptation_harp_dataset}
\end{figure*}

\begin{figure}[t]
\begin{center}
\includegraphics[width=\linewidth]{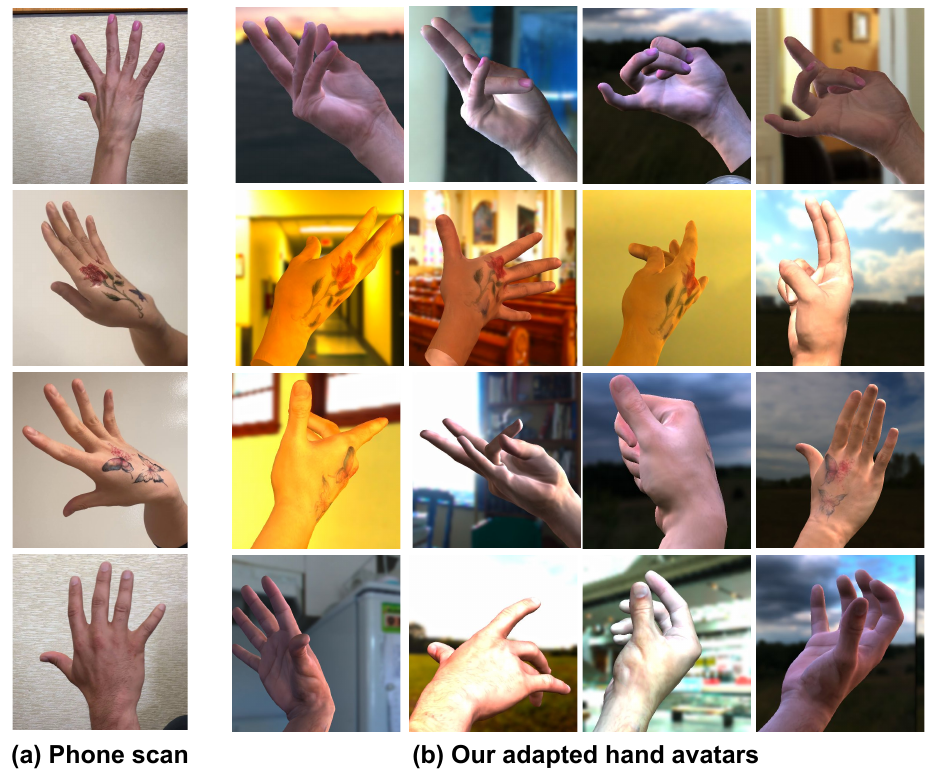}
\end{center}
\vspace*{-7mm}
\caption{
Our adapted 3D hand avatar with the Phong reflection model and environment maps~\cite{gardner2017learning,hold2019deep}.
}
\vspace*{-3mm}
\label{fig:avatar_phong}
\end{figure}

\subsection{3D hand avatars}
Fig.~\ref{fig:compare_adaptation_harp_dataset} shows that our 3D hand avatar achieves sharper textures than HARP~\cite{karunratanakul2023harp}.
Handy~\cite{potamias2023handy} fails to produce authentic results, consistent with Fig.~\textcolor{red}{10} and ~\textcolor{red}{11} of the main manuscript.
Fig.~\ref{fig:avatar_phong} additionally shows our adapted 3D hand avatar, rendered with Phong reflection model and environment maps, as in Fig.~\textcolor{red}{1} (b) of the main manuscript.
To this end, given an environment map, we first do preconvolution to map the illumination in the environment map to diffuse and specular lighting representation similar to~\cite{pandey2021total}.
Then, the final texture is obtained by combining the diffuse and specular representation with our adapted texture (optimized texture of Sec.~\textcolor{red}{5.4} of the main manuscript) according to the normal map from 3D mesh and view direction.
The 3D poses of (b) are from the tracked results from a different subject of our studio data, which shows that our hand avatars can be driven with novel poses.
The results are not photorealistic due to the limitation of the Phong reflection model, but they show the potential of our hand avatar, which can be combined with future relightable hand models~\cite{chen2024urhand}.

%% file: src_suppl/more_ablation_studies.tex
\begin{figure}[t]
\begin{center}
\includegraphics[width=\linewidth]{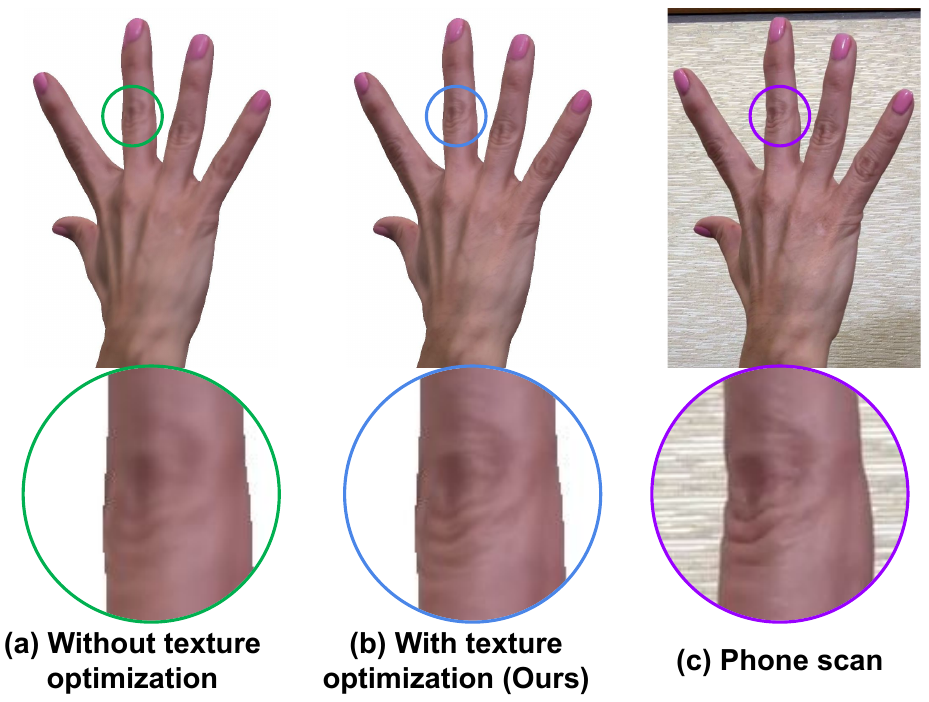}
\end{center}
\vspace*{-7mm}
\caption{
The effectiveness of the texture optimization during our phone adaptation.
}
\vspace*{-3mm}
\label{fig:texture_optimization}
\end{figure}

\section{More ablation studies}\label{sec:ablation_study_suppl}

\subsection{Effectiveness of the texture optimization}
Fig.~\ref{fig:texture_optimization} shows that our texture optimization, described in Sec.~\textcolor{red}{5.4} of the main manuscript, further enhances the photorealism of the texture.

\subsection{Effectiveness of the TV regularizer during the adaptation}
\begin{figure}[t]
\begin{center}
\includegraphics[width=\linewidth]{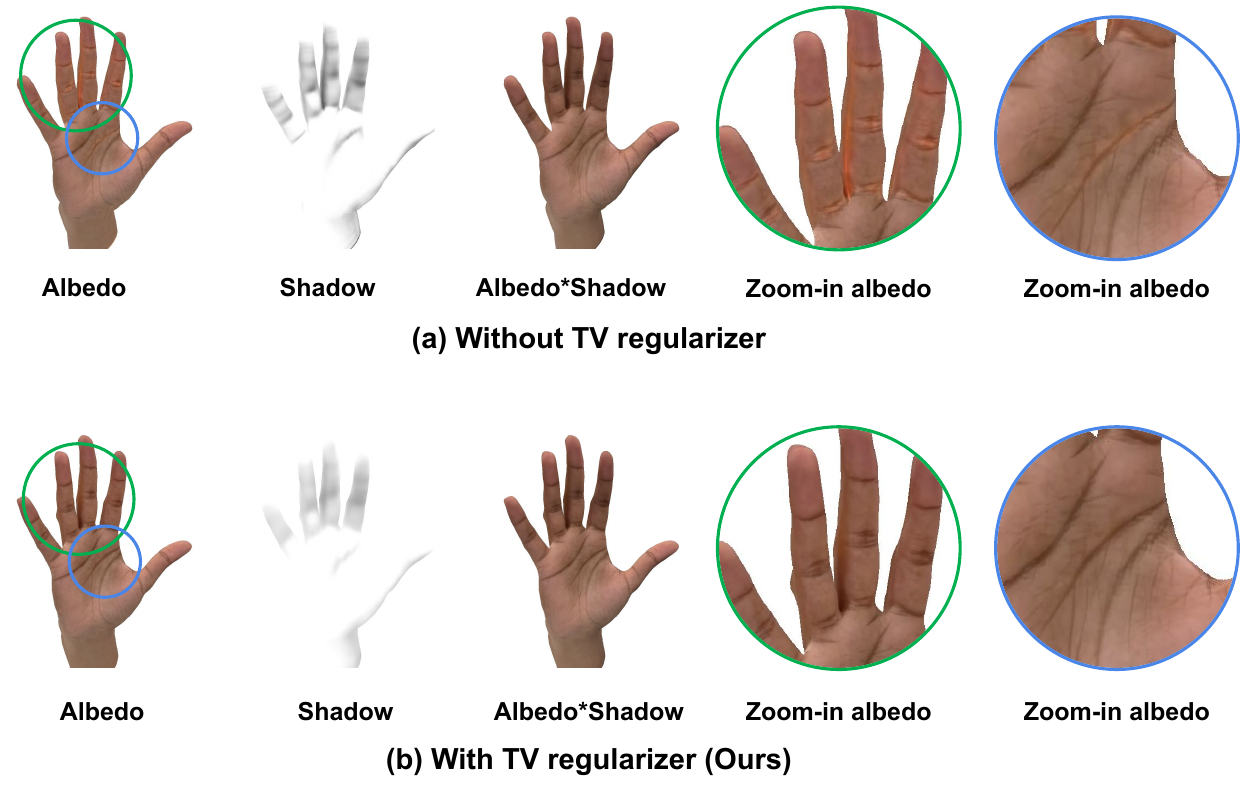}
\end{center}
\vspace*{-7mm}
\caption{
The effectiveness of the total variation (TV) regularizer to the ShadowNet.
}
\vspace*{-3mm}
\label{fig:shadownet_tv_reg}
\end{figure}

Fig.~\ref{fig:shadownet_tv_reg} shows the effectiveness of the total variation (TV) regularizer to our ShadowNet.
Without the TV regularizer, ShadowNet tried to consider all darkness differences between 1) albedo+shadow and 2) captured images as shadows.
As a result, local sharp textures, including wrinkles are considered shadows.
As described in Sec.~\textcolor{red}{5.3} of the main manuscript, by applying the TV regularizer to the ShadowNet, we can prevent such undesired shadows.

\subsection{Extension of DHM to the universal case vs. UHM.}

\begin{figure}[t]
\begin{center}
\includegraphics[width=\linewidth]{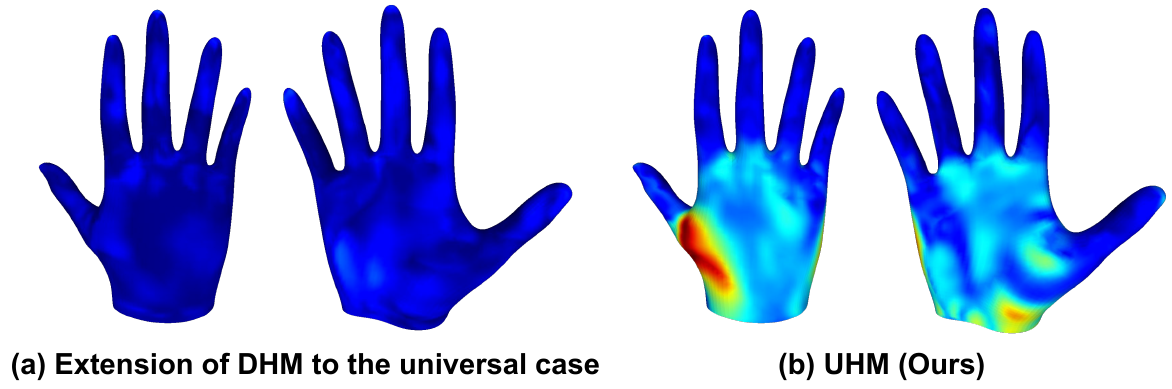}
\end{center}
\vspace*{-7mm}
\caption{
Comparison of the standard deviation of ID-dependent vertex corrective $\Delta\bar{\bm{V}}^\text{id}$.
The correctives for the standard deviation computation are obtained by 1) randomly sampling 512 ID code $\mathbf{z}^\text{id}$ from the normal Gaussian and 2) passing them to pre-trained IDDecoder.
}
\vspace*{-3mm}
\label{fig:id_corr_std}
\end{figure}

Fig.~\ref{fig:id_corr_std} shows that when performing the tracking and modeling at the same time, special considerations are necessary for universal hand modeling.
We choose DHM~\cite{moon2020deephandmesh}, a high-fidelity personalized 3D hand model, as a comparison target because it has a similar training pipeline that performs the tracking and modeling at the same time as ours.
One critical difference between DHM and our UHM is that DHM is a personalized 3D hand model, which does not learn the ID space and cannot generalize to novel IDs.
For systems that perform the tracking and modeling at the same time, one major difficulty of universal hand modeling is disentangling ID and pose information as all supervision targets, such as 3D joint coordinates, 3D scans, masks, and images, are entangled representations of ID and pose.
We effectively achieve the disentanglement by calculating loss functions using two types of 3D meshes: one from both correctives and the other only from the ID-dependent correctives, as described in Sec.~\textcolor{red}{4} of the main manuscript.
During the training, the ID-dependent correctives of all frames that belong to the same ID are from the same inputs (\textit{i.e.}, 3D joint coordinates and depth maps of the neutral pose, as described in \textbf{IDEncoder and IDDecoder.} of Sec.~\textcolor{red}{3.2} of the main manuscript).
Therefore, supervising 3D meshes that are only from the ID-dependent correctives can make IDDecoder formulate meaningful ID space (Fig.~\ref{fig:id_corr_std} (b)) without being affected by the pose-and-ID-dependent correctives, which naturally achieves the disentanglement of the ID and pose.
On the other hand, without the supervision of the 3D meshes that are only from the ID-dependent correctives like DHM, the model cannot disentangle ID and pose, which results in meaningless ID space (Fig.~\ref{fig:id_corr_std} (a)).
Such disentanglement is especially challenging for systems that perform the tracking and modeling at the same time because previous separate pipeline~\cite{romero2017embodied,li2022nimble,potamias2023handy} can perform tracking for each ID, which can naturally provide assets that only have ID information without pose by canceling pose from the tracked meshes.

%% file: src_suppl/uhm_network_architectures_loss_functions.tex
\section{UHM network architectures and regularizers}\label{sec:network_architecture_and_regularizers_suppl}

\subsection{Network architectures}
We describe detailed network architectures of UHM, briefly described in Sec.~\textcolor{red}{3.2} of the main manuscript.

\noindent\textbf{IDEncoder.}
IDEncoder outputs ID code $\mathbf{z}^\text{id} \in \mathbb{R}^{32}$ from a pair of a depth map and 3D joint coordinates of each training subject.
To prepare the inputs of the IDEncoder, we first select a single pair of a 3D scan and 3D joint coordinates for each subject.
Hence, there are subjects number of (3D scan and 3D joint) pairs.
As the IDEncoder should capture only ID-related information, the poses of the inputs of the IDEncoder should be normalized.
To this end, we take the pairs from the first frame of the captures as the poses at the first frames are close to zero poses, which we call \emph{neutral poses}.
Then, we rigidly align the selected 3D scans and 3D joint coordinates to a reference coordinate system and render depth maps from the aligned 3D scans from the front and back views.
In this way, we can further normalize views, which exist and are hard to be normalized in images.

ResNet-18~\cite{he2016deep} takes two-view depth maps of neutral pose for each subject.
Please note that IDEncoder always takes the same inputs for the same subject during the training.
Hence, the size of the mini-batch is $2N_\text{s}$, where $N_\text{s}$ is the number of unique subjects in the mini-batch of PoseEncoder.
The ResNet-18 is initialized with ImageNet~\cite{russakovsky2015imagenet} classification, and we discard fully connected layers.
The output of ResNet-18 is a 512-dimensional feature vector.
We reshape the feature vector to a 1024-dimensional one, which represents a multi-view feature for each subject.
The multi-view feature is concatenated with the 3D joint coordinate of a neutral pose and passed to two fully connected layers, which produce the id code $\mathbf{z}^\text{id}$ using the reparameterization trick~\cite{kingma2014vae}.
The two fully connected layers consist of 512 hidden units and an intermediate ReLU activation function.

\noindent\textbf{IDDecoder.}
IDDecoder takes the ID code $\mathbf{z}^\text{id}$ and outputs ID-dependent skeleton correctives $\Delta\bar{\bm{J}}^\text{id}$ and ID-dependent vertex correctives $\Delta\bar{\bm{V}}^\text{id}$.
The IDDecoder consists of two fully connected layers with a ReLU activation function for the non-linearity.
The hidden size of the fully connected layers is set to 512.
$\Delta\bar{\bm{J}}^\text{id}$ should not replicate any changes, which can be replicated by 3D joint rotations.
In other words, 3D hands should be in the ``zero pose" after applying $\Delta\bar{\bm{J}}^\text{id}$ to the template mesh.
Hence, except for child joints of the wrist, we enable only 1 degree of freedom (DoF) of $\Delta\bar{\bm{J}}^\text{id}$ to restrict it to only affect the lengths of fingers.
In this way, the learned ID space is not mixed with the pose.

\noindent\textbf{PoseEncoder.}
PoseEncoder outputs 6D rotation~\cite{zhou2019continuity} of joints $\bm{\Theta}$ from a pair of a single RGB image and 3D joint coordinates of arbitrary poses and identities.
The 3D global rotation and translation are obtained by rigidly aligning wrist and four finger root joints (except the thumb root joint) to the target 3D joint coordinates.
Unlike IDEncoder's inputs consist of a single pair of each subject, PoseEncoder's inputs are from any poses and subjects.
Our PoseEncoder has a similar network architecture as Pose2Pose~\cite{moon2022hand4whole}.
The ResNet-50 of Pose2Pose is initialized with ImageNet~\cite{russakovsky2015imagenet} classification, and the remaining parts are randomly initialized.

\noindent\textbf{PoseDecoder.}
PoseDecoder outputs pose-and-ID-dependent vertex corrective $\Delta\bar{\bm{V}}^\text{pose}$ in a sparse way using local joint clusters for better generalization to unseen poses following STAR~\cite{osman2020star}.
To this end, we make $J$ number of local joint clusters, where each cluster consists of 6D rotations of a joint, its parent joint, and a child joint.
$J$ denotes the number of joints.
We additionally concatenate the ID code for each cluster.
Hence, each local joint cluster has the dimension of $\mathbb{R}^{18+32}$, where 18 and 32 represent three 6D rotations and the dimension of the ID code, respectively.
Please note that we pass 6D rotations after masking invalid DoFs and root rotation to zero.
Then, the local joint clusters are passed to two separable convolutions with an intermediate ReLU activation function for a non-linearity.
The hidden size of the separable convolution is set to 256.
The output of the separable convolutions of each local joint cluster has the dimension of $\mathbb{R}^{V \times 3}$.
$V$ denotes the number of vertices of our template mesh.
Formally, we denote the above process by $\bm{F}_j=f(\theta_j, \theta_{p(j)}, \theta_{c(j)}, \mathbf{z}^\text{id})$, where $\bm{F}_j$ denotes the output of the separable convolution of $j$th local joint cluster.
$p(j)$ and $c(j)$ denote parent child joint of $j$th joint, respectively.
The final pose-and-ID-dependent vertex corrective $\Delta\bar{\bm{V}}^\text{pose}$ is obtained by $\Delta\bar{\bm{V}}^\text{pose}=\sum_j \bm{\Phi}_j (f(\theta_j, \theta_{p(j)}, \theta_{c(j)}, \mathbf{z}^\text{id}) - f(\mathbf{0}, \mathbf{0}, \mathbf{0}, \mathbf{z}^\text{id}))$.
$\bm{\Phi} \in \mathbb{R}^{V \times J}$ is a mask, which introduces sparsity.
It is initialized with a geodesic distance between the $v$th vertex and $j$th joint in the template mesh.  
We subtract the output of the separable convolution from the zero pose to prevent pose-and-ID-dependent vertex corrective $\Delta\bar{\bm{V}}^\text{pose}$ from replicating only ID-dependent geometry, which should be replicated by ID-dependent vertex corrective $\Delta\bar{\bm{V}}^\text{id}$.

\subsection{Loss functions for the tracking and modeling}

Our UHM is trained in an end-to-end manner by minimizing $L$, defined as below:
\begin{equation}
\begin{aligned}
L = & L_\text{pose} + 10 L_\text{p2p} + 0.1 L_\text{mask} + 0.1 L_\text{img} \\ 
& + 0.01 L_{\bm{\Theta}} + 0.001 L_{\mathbf{z}^\text{id}} + 1000 L_{\Delta\bar{\bm{V}}^\text{id}} + 10 L_{\Delta\bar{\bm{V}}^\text{pose}} \\ 
& + 75000 L_\text{lap} + 0.001 L_{\bm{\Phi}} + 0.1 L_\text{vol} + 0.1 L_\text{cut},
\end{aligned}
\end{equation}
where $L_\text{pose}$, $L_\text{p2p}$, $L_\text{mask}$, and $L_\text{img}$ are described in the Sec.~\textcolor{red}{4} of the main manuscript.
The remaining loss functions are regularizers, described below.

First, we minimize $L_{\bm{\Theta}}$, a squared $L2$ norm of $\bm{\Theta}$ after converting it to an axis-angle representation, to prevent extreme rotations.
Second, we minimize $L_{\mathbf{z}^\text{id}}$, a KL divergence between $\mathbf{z}^\text{id}$ and the normal Gaussian distribution.
In this way, we can make the ID latent space follow the Gaussian distribution, necessary for sampling novel ID from a known (\textit{e.g.}, Gaussian) distribution.
Third, we minimize $L_{\Delta\bar{\bm{V}}^\text{id}}$ and $L_{\Delta\bar{\bm{V}}^\text{pose}}$, a squared $L2$ norm of the tangential component of $\Delta\bar{\bm{V}}^\text{id}$ and $\Delta\bar{\bm{V}}^\text{pose}$, respectively.
They prevent the vertex correctives from overwhelming the ID-dependent skeleton corrective $\Delta\bar{\bm{J}}^\text{id}$.
To be more specific, we encourage the finger lengths to be adjusted mainly by $\Delta\bar{\bm{J}}^\text{id}$, not by the vertex correctives.
Fourth, we minimize $L_\text{lap}$, the Laplacian regularizer for smooth surface.
Like $L_\text{p2p}$ and $L_\text{mask}$, we compute two types of this regularizer from 1) both correctives and 2) only ID-dependent corrective to learn meaningful ID space.
Fifth, we minimize $L_{\bm{\Phi}}$, a $L1$ norm of $\mathrm{ReLU}(\bm{\Phi})$, following STAR~\cite{osman2020star}.
In this way, we can encourage sparsity of the $\Delta\bar{\bm{V}}^\text{pose}$, beneficial for the generalizability to unseen 3D poses.
Sixth, we minimize $L_\text{vol}$, a volume-preserving regularizer.
It first pre-calculates the radius of spheres for each finger in the zero pose space only with $\Delta\bar{\bm{V}}^\text{id}$ without $\Delta\bar{\bm{V}}^\text{pose}$.
Then, $L_\text{vol}$ is the difference between 1) the distance from vertices to sphere radius and 2) the radius of the sphere if the distance shorter than the radius.
It encourages our UHM to preserve the minimal volume of each finger, where the minimal values are calculated in the zero pose space with $\Delta\bar{\bm{V}}^\text{id}$.
Finally, we minimize $L_\text{cut}$ for a flat cut around the forearm.
To this end, we make a virtual vertex at the center of the cut and make virtual triangles using the virtual vertex and pairs of two connected vertices at the cut.
$L_\text{cut}$ is a $L1$ distance between dot products of all those virtual triangles.
In this way, we can encourage all vertices at the cut to be on the same plane, which results in a flat cut.

%% file: src_suppl/details_adaptation_to_a_phone_scan.tex
\begin{figure*}[t]
\begin{center}
\includegraphics[width=\linewidth]{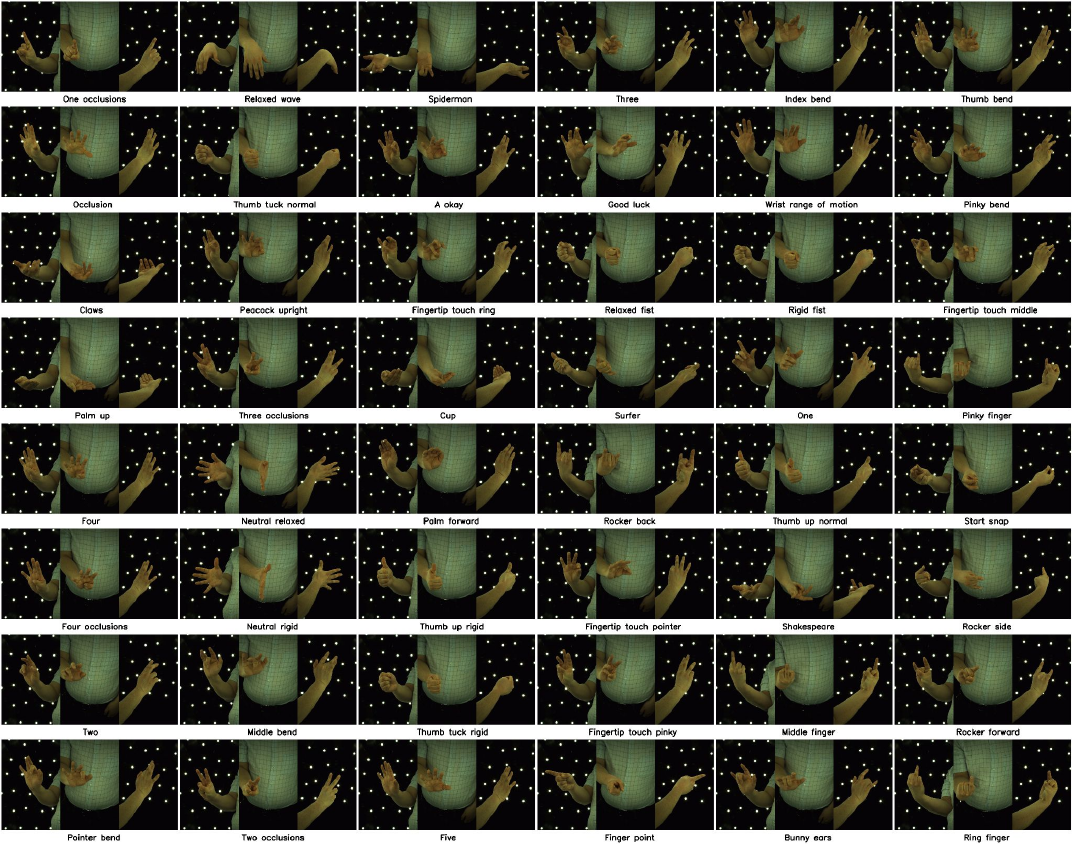}
\end{center}
\vspace*{-7mm}
\caption{
Examples of poses of the training set of our studio dataset.
}
\vspace*{-3mm}
\label{fig:training_set_poses}
\end{figure*}

\begin{figure*}[t]
\begin{center}
\includegraphics[width=\linewidth]{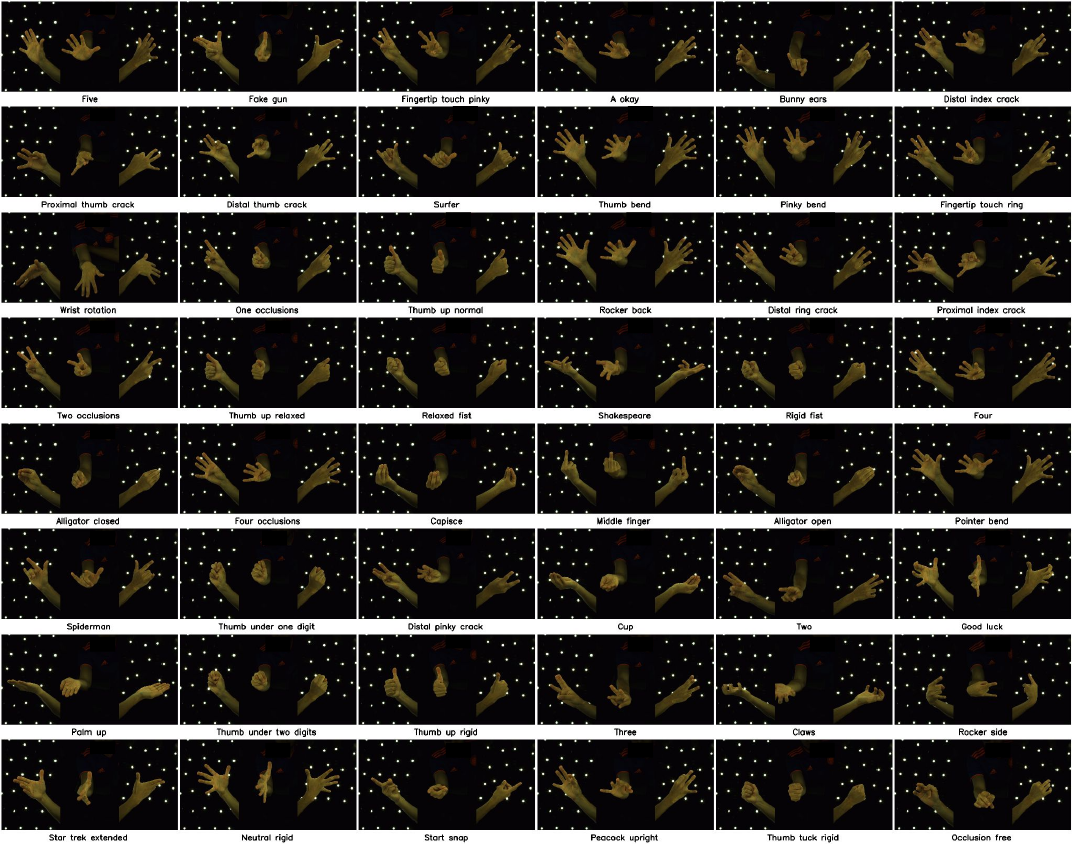}
\end{center}
\vspace*{-7mm}
\caption{
Examples of poses of the testing set our studio dataset.
}
\vspace*{-3mm}
\label{fig:testing_set_poses}
\end{figure*}

\section{Details of adaptation to a phone scan}\label{sec:adaptation_suppl}

We provide detailed descriptions of our adaptation stage, described in Sec.~\textcolor{red}{5} of the main manuscript.

\subsection{Geometry fitting}\label{subsec:adaptation_geo_fitting_suppl}

For the geometry fitting, we designed PoseNet, which has a similar network architecture to Pose2Pose~\cite{moon2022hand4whole} with minor modifications.
The PoseNet outputs 3D global rotation, 3D pose $\bm{\Theta}$, pose code, and 3D global translation of UHM from an image, a depth map, a mask, and 2D joint coordinates, where the inputs are obtained from Sec.~\textcolor{red}{5.1} of the main manuscript.
The pose code is a latent vector of a pre-trained VAE, which embeds plausible hand pose space similar to V-Poser~\cite{pavlakos2019expressive}.
The VAE is pre-trained on our capture studio dataset and fixed during the adaptation stage.

We randomly initialize PoseNet before the training.
The PoseNet is used for the pose tracking, a similar spirit of neural annotators~\cite{moon2022neuralannot}.
In addition to the outputs of the network, we optimize ID code $\mathbf{z}^\text{id}$, shared across all frames as all frames are from the same person.
With the outputs of the network and the ID code, we obtain 3D mesh using pre-trained decoders of UHM, used to unwrap images to the UV space.

The PoseNet is trained in a self-supervised way by being trained with the inputs of the network (\textit{i.e.}, 2D joint coordinates, a depth map, and a mask).
During the training, we fixed pre-trained decoders of UHM.
For the kinematic-level personalization (\textit{e.g.}, bone lengths), we minimize $L1$ distance between projected 2D hand joint coordinates and the target.
Also, for the surface-level personalization (\textit{e.g.}, thickness of hand surface), we minimize $L1$ distance between the rendered mask and the target.
We additionally minimize the $L1$ distance between the rendered depth map and the target to address the depth and scale ambiguity.
Finally, we minimize the $L1$ distance between 3D joint coordinates from the pose code and MANO parameters, where the MANO parameters are from an off-the-shelf regressor~\cite{moon2023interwild}.
In this way, we can address the depth ambiguity of the 2D keypoints.
Then, 3D joint angles and 3D mesh from the pose code are used to supervise those from the 3D pose $\bm{\Theta}$.

\subsection{ShadowNet}

We first tile 3D global rotation, 3D pose $\bm{\Theta}$, and ID code $\mathbf{z}^\text{id}$ to all texels in the UV space.
In other words, all texels have the same concatenated 3D global rotation, 3D pose, and ID code.
Then, we compute the dot product between 1) the normal vector of each vertex and 2) a vector from the camera to each vertex, which becomes a viewpoint feature for each vertex.
We warp the per-vertex viewpoint feature to the UV space and concatenate it with the prepared tiled texels, which become the input of our ShadowNet.
Given a fixed environment during the phone scan, all inputs of our ShadowNet can determine casted shadow.
To distinguish each texel with its own semantic meaning, we add a learnable positional encoding to each texel and pass it to ShadowNet.
To enlarge the size of the receptive field effectively, we start from 32 downsampled UV space compared to that of our UV textures.

The ShadowNet first converts the input to a 256-dimensional feature map with a convolutional layer.
Then, for each resolution, we apply one convolutional layer, followed by group normalization~\cite{wu2018group} and SiLU activation function~\cite{elfwing2018sigmoid}.
We used the nearest neighbor for the upsampling.
After three times of upsampling, we apply bilinear upsampling four times and the sigmoid activation function to normalize the values of the shadow from 0 to 1.

%% file: src_suppl/our_datasets.tex
\section{Our datasets}\label{sec:our_datasets_suppl}

We provide detailed descriptions of our two types of datasets.

\subsection{Studio dataset}
Our capture studio has 170 calibrated and synchronized cameras.
All cameras lie on the front, side, and top hemispheres of the hand and are placed at a distance of about one meter from it.
Images are captured with 4096 $\times$ 2668 pixels at 30 frames per second.
We pre-processed the raw video data by performing 2D keypoint detection~\cite{sun2019deep} and 3D scan~\cite{guo2019relightables}. 
The keypoint detector is trained on our held-out human annotation dataset, which includes 900K images with 3D rotation center coordinates of hand joints, where our manual annotation tool is similar to that of Moon~\etal~\cite{Moon_2020_ECCV_InterHand2.6M}.
The predicted 2D keypoints of each view were triangulated with RANSAC to robustly obtain the groundtruth (GT) 3D hand joint coordinates. 
The combination of 2D keypoint detector and triangulation, used to obtain GT 3D hand joint coordinates, achieves a 1.71 mm error on our held-out human-annotated test set, which is quite low.
Fig.~\ref{fig:training_set_poses} and ~\ref{fig:testing_set_poses} show pose examples of the training and testing sets of our studio dataset, respectively.

\begin{figure*}[t]
\begin{center}
\includegraphics[width=\linewidth]{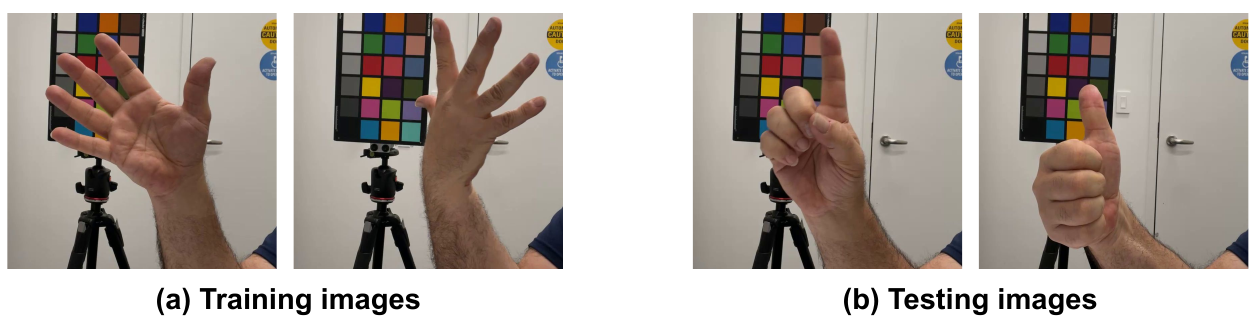}
\end{center}
\vspace*{-7mm}
\caption{
Examples of our phone scan dataset.
}
\vspace*{-3mm}
\label{fig:phone_scan_exmaples}
\end{figure*}

\subsection{Phone scan dataset}
Fig.~\ref{fig:phone_scan_exmaples} shows examples of our phone scan dataset.
The training set mainly consists of simple poses, where the 3D global rotation of the hand mainly changes, and the 3D pose and 3D translation of the hand remain almost static.
The testing set consists of diverse poses, such as a fist and thumb-up.

%% file: src_suppl/experiment_details.tex
\section{Experiment details}\label{sec:experiment_details_suppl}

\subsection{Fitting for Sec.~\textcolor{red}{6.2}}

For the comparisons in Fig.~\textcolor{red}{9} and Tab.~\textcolor{red}{1} of the main manuscript, we used 3D joint coordinates and 3D scans as target data for the fitting, the most typical setting of the tracking.
For the fitting, we minimized 1) $L1$ distance between output and target 3D joint coordinates, 2) the P2S distance from 3D scans, and 3) $L2$ regularizers to the parameters.
The $L2$ regularizer is introduced to prevent extreme meshes.
Each loss term is weighted by 1, 10, and 0.001.
As each 3D hand model has slightly different 3D joint locations despite the same semantic meaning, we do not report 3D joint error following ~\cite{romero2017embodied,li2022nimble,corona2022lisa}.
For the same reason, we turned off the 3D joint loss during the fitting after enough iterations.

For the comparison in Tab.~\textcolor{red}{2} of the main manuscript, we followed the evaluation protocol of LISA~\cite{corona2022lisa}.
Specifically, we pre-define various numbers of available viewpoints and fit 3D pose and ID code to 2D joint coordinates of those viewpoints.
Due to the depth ambiguity from 2D supervisions from a few viewpoints, we used the pose prior, used in our adaptation stage of Sec.~\ref{subsec:adaptation_geo_fitting_suppl}, as LISA also used geometry prior from large-scale data.
As their codes are not publicly available, we brought their numbers from their paper.

\subsection{Handy fitting for Sec.~\textcolor{red}{6.3}}

We use the same PoseNet and loss functions of ours, described in Sec.~\ref{sec:adaptation_suppl}, except for one thing: we used VGG loss function~\cite{ledig2017photo} on the rendered image, while we used LPIPS~\cite{zhang2018unreasonable} on the rendered image for the Handy texture fitting following their paper.
The latent code of the Handy's texture is shared across all frames and is optimizable.

\begin{figure}[t]
\begin{center}
\includegraphics[width=\linewidth]{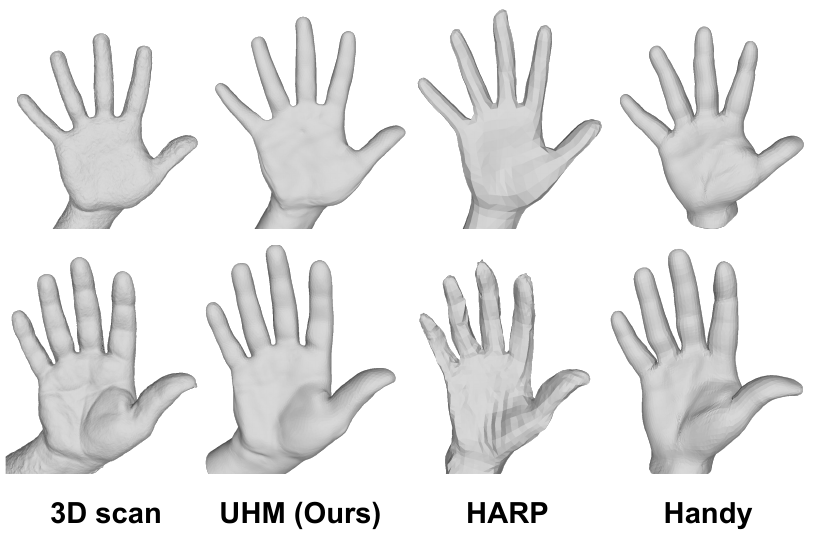}
\end{center}
\vspace*{-7mm}
\caption{
Comparison of 3D scan and 3D meshes from various hand adaptation pipelines.
}
\vspace*{-3mm}
\label{fig:adaptation_compare_3d}
\end{figure}

\subsection{P2S calculation of Tab.~\textcolor{red}{3}}
We calculated the P2S errors of the adapted 3D hand avatar in Tab.~\textcolor{red}{3} of the main manuscript.
To this end, we first selected a frame with the neutral pose of studio capture of the four subjects, where the four subjects have co-captured studio and phone scan data.
From the studio data, we have 3D joint coordinates and 3D scan of the neutral pose.
Then, we optimize 3D pose and translation of each adapted avatars by minimizing $L1$ 3D joint distance and point-to-point loss function from the studio data, described in Sec.~\textcolor{red}{4} of the main manuscript.
During the optimization, we fix ID-related information, such as ID code $\mathbf{z}^\text{id}$ of ours.
The P2S errors are calculated between the optimized meshes of each avatar and 3D scan from our studio data.
Fig.~\ref{fig:adaptation_compare_3d} visualizes the optimized mesh and 3D scan.
For UHM and HARP, we excluded the vertices on the forearm when calculating the 3D errors as they are too unconstrained.

\subsection{HARP dataset}
As the HARP dataset does not provide depth maps, we do not use the depth map loss function in our pipeline.
We used Mediapipe~\cite{mediapipe} to obtain 2D hand joint coordinates and used RVM~\cite{lin2022robust} to obtain foreground masks.
All remaining things are the same as what is described in Sec.~\ref{sec:adaptation_suppl} for the experiments on the HARP dataset.
Ours, HARP, and Handy are equally fitted to the same sequences and are evaluated with the same metrics.

%% file: src_suppl/failure_cases.tex
\begin{figure}[t]
\begin{center}
\includegraphics[width=\linewidth]{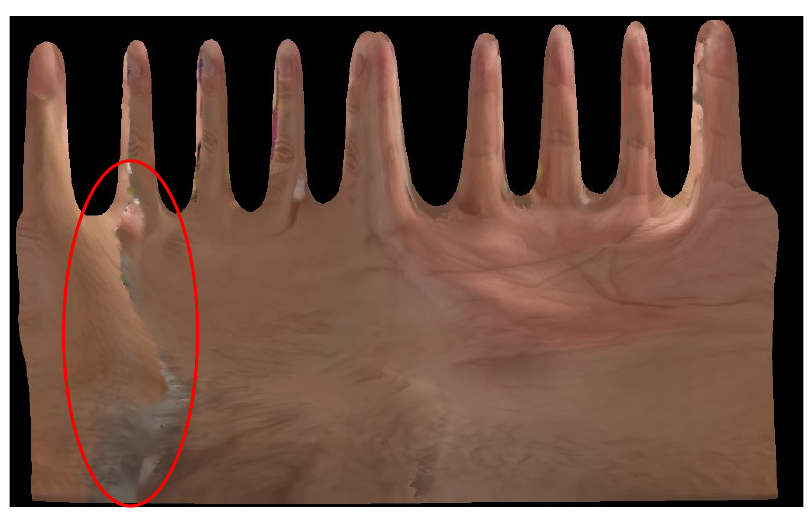}
\end{center}
\vspace*{-7mm}
\caption{
Our optimized texture after removing shadow with the ShadowNet.
The highlighted area has an evident artifact.
}
\vspace*{-3mm}
\label{fig:failure_case_unwrap}
\end{figure}

\begin{figure}[t]
\begin{center}
\includegraphics[width=\linewidth]{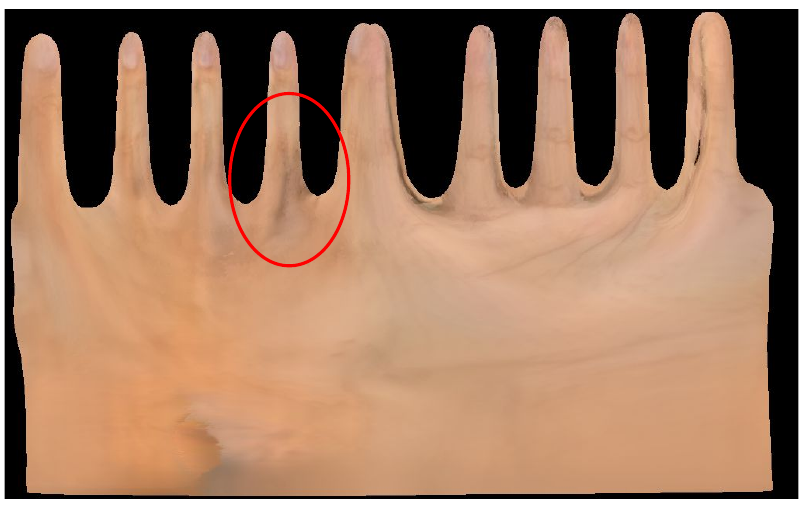}
\end{center}
\vspace*{-7mm}
\caption{
Our optimized texture after removing shadow with the ShadowNet.
The highlighted area has an evident artifact.
}
\vspace*{-3mm}
\label{fig:failure_case_shadownet}
\end{figure}

\section{Failure cases}\label{sec:failure_cases_suppl}

\noindent\textbf{Geometry fitting.}
We found that our geometry fitting pipeline (Sec.~\textcolor{red}{5.2} of the main manuscript) sometimes suffers from a surface-level misalignment.
In the geometry fitting stage, dense supervisions, such as DensePose~\cite{guler2018densepose} of the 3D human body, are not available.
Such a lack of dense supervision makes our 3D geometry suffer from surface-level misalignment despite the accurate keypoint-level alignment.
Although the image loss during the texture optimization (Sec.~\textcolor{red}{5.2} of the main manuscript) provides the dense supervision, its initial texture is from the geometry fitting (Sec.~\textcolor{red}{5.2} of the main manuscript), which can suffer from the surface-level misalignment.

\noindent\textbf{Texture unwrapping.}
Fig.~\ref{fig:failure_case_unwrap} shows a failure case happens in the texture unwrapping.
There is an evident vertical artifact along the left part of the figure.
The reason for such artifacts is that during the phone capture, the subject exposes the left and right parts of the vertical line with very different poses at different time steps.
Hence, pose-dependent skin color changes and view-dependent shading of those left and right parts become very different, which results in different colors and an evident vertical line between the left and right parts.
We tried to smooth such a region; however, it was not enough as the color difference is too big.

\noindent\textbf{ShadowNet.}
Fig.~\ref{fig:failure_case_shadownet} shows a failure case of our ShadowNet.
Although most of the shadow is removed, the highlighted area still has a small amount of shadow.
The remaining shadow is especially evident as the skin color of this subject is bright.
We think the reason for the remaining shadow is the regularizers to the ShadowNet to prevent it from considering black tattoos as shadows.
Also, its capability is not guaranteed for \emph{smooth} black tattoos and black fingernail polish.
Due to the ambiguity of the intrinsic decomposition, it might perform badly in low-light conditions; we think this limitation applies to all current methods.